\documentclass[10pt,twocolumn,letterpaper]{article}

\usepackage[pagenumbers]{cvpr} % To force page numbers, e.g. for an arXiv version

\usepackage{multirow}
\usepackage{graphicx}

\PassOptionsToPackage{table}{xcolor} 

\usepackage{multicol}
\usepackage{multirow}
\usepackage{color,soul}
\usepackage{makecell} 
\usepackage[table,xcdraw]{xcolor}  
\usepackage{colortbl}    
\usepackage[accsupp]{axessibility}
\usepackage{algorithmic}
\usepackage{algorithm}
\definecolor{cvprblue}{rgb}{0.21,0.49,0.74}
\usepackage[pagebackref,breaklinks,colorlinks,allcolors=cvprblue]{hyperref}

\title{Physically Inspired Gaussian Splatting for HDR Novel View Synthesis}

\author{Huimin Zeng\textsuperscript{1} \qquad  Yue Bai\textsuperscript{1}  \qquad  Hailing Wang\textsuperscript{1}  \qquad  Yun Fu\textsuperscript{1,2}\\
\textsuperscript{1} Department of Electrical and Computer Engineering, Northeastern University \\
\textsuperscript{\rm 2}Khoury College of Computer Science, Northeastern University\\
{\tt\small  \{zeng.huim, bai.yue, wang.haili, y.fu\}@northeastern.edu}
}

\begin{document}
\maketitle
\begin{abstract}
High dynamic range novel view synthesis (HDR-NVS) reconstructs scenes with dynamic details by fusing multi-exposure low dynamic range (LDR) views, yet it struggles to capture ambient illumination-dependent appearance. Implicitly supervising HDR content by constraining tone-mapped results fails in correcting abnormal HDR values, and results in limited gradients for Gaussians in under/over-exposed regions. To this end,  we introduce PhysHDR-GS,  a physically inspired HDR-NVS framework that models scene appearance via intrinsic reflectance and adjustable ambient illumination.   PhysHDR-GS employs a complementary image-exposure (IE) branch and Gaussian-illumination (GI) branch to faithfully reproduce standard camera observations and capture illumination-dependent appearance changes, respectively. During training, the proposed cross-branch HDR consistency loss provides explicit supervision for HDR content, while an illumination-guided gradient scaling strategy mitigates exposure-biased gradient starvation and reduces under-densified representations.  Experimental results across realistic and synthetic datasets demonstrate our superiority in reconstructing HDR details (\eg, a PSNR gain of \textbf{2.04 dB} over HDR-GS), while maintaining real-time rendering speed (up to \textbf{76 FPS}).   Code and models are available at \url{https://huimin-zeng.github.io/PhysHDR-GS/}.
\end{abstract}

 \section{Introduction}\label{sec:intro}
Novel view synthesis (NVS)~\cite{3dgs,nerf,lu2024scaffold,paliwal2024coherentgs,xu2024wild} reconstructs a 3D scene from sparse image sequences and renders high-quality novel views, demonstrating wide applications in scenarios such as autonomous driving~\cite{altillawi2025npbg,ma2025novel}  and AR/VR~\cite{muller2024multidiff,fu2023auto}. However, due to limited dynamic range of standard sensors, captured sequences are unable to faithfully reflect varying illumination in real-world scenes, resulting in missing details in reconstructed highlights and shadows. High dynamic range novel view synthesis (HDR-NVS)~\cite{huang2022hdr,jun2022hdr,huang2024ltm,singh2024hdrsplat, jin2024lighting,li2024chaos,li2024chaos,mildenhall2022nerf,cai2024hdr,wu2024hdrgs} addresses this by leveraging multi-exposures fusion (MEF) to combine complementary information from low dynamic range (LDR) images captured with different exposures. Recent advances from Neural Radiance Field (NeRF)~\cite{nerf} to 3D Gaussian Splatting (3DGS)~\cite{3dgs} largely accelerate HDR-NVS task, enabling high-quality and real-time rendering.

\begin{figure}[t]
    \centering
    \includegraphics[width=0.83\linewidth, clip, trim=10pt 18pt 0 0]{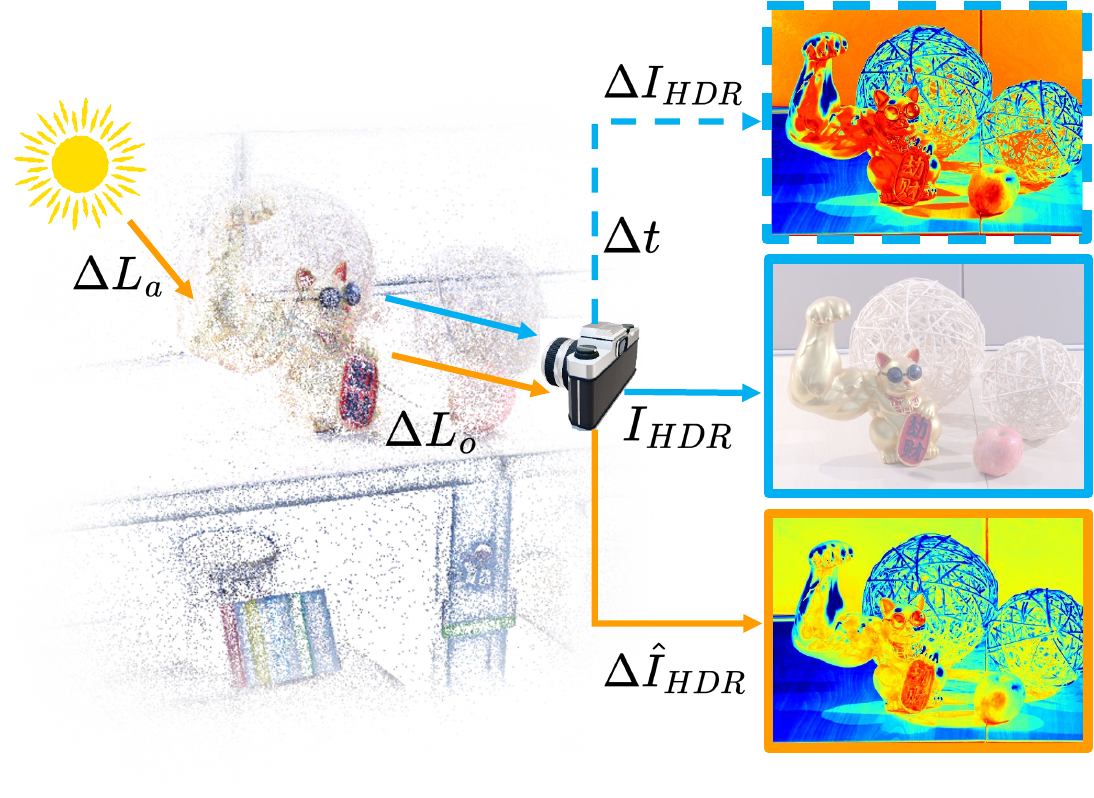}\vspace{-2pt}
    \caption{Variation of camera exposure $\Delta t$ and ambient illumination $\Delta L_a$ scales the HDR signal in different ways: $\Delta t$ causes a global change $\Delta I_{HDR}$, while $\Delta L_a$ induces local changes $\Delta \hat{I}_{HDR}$ (\eg, nameplate of the luckycat) via lighting-conditioned radiance variation $\Delta L_o$.  Their different response patterns reveal complementary ways of modeling dynamic-range details.} 
    \label{fig:teaser}
\end{figure}

Despite high dynamic details introduced by MEF, HDR-NVS still faces several challenges: \textit{(1) appearance entanglement:} the appearance of objects is jointly determined by material properties and environmental conditions (\eg, direct/indirect illumination). Simply scaling sensor shutter time (\ie, exposure $t$) cannot disentangle these factors and reflect illumination-dependent appearance changes. As shown in \cref{fig:teaser}, a change in exposure $\Delta t$ mainly causes global intensity change $\Delta I_{HDR}$, while ambient illumination change $\Delta L_a$  induces local change $\Delta \hat{I}_{HDR}$ (\eg, the luckycat nameplate) via radiance variation $\Delta L_o$; \textit{(2) implicit HDR supervision:}  HDR ground truth (GT) is typically unavailable. Supervision of reconstructed HDR view is therefore implicitly conducted by constraining LDR views tonemapped from HDR views. Since tone mapping compresses dynamic ranges, abnormal or saturated HDR values cannot be reflected in tonemapped results and be effectively constrained; \textit{(3) exposure-biased gradient starvation:} tone mapping curve typically yields small slopes at the extremes. Therefore,  Gaussians covering under/over-exposed regions accumulate much smaller gradients than those at normal exposures (see \cref{sec:gradient}). As a result, they struggle to meet densification thresholds, leading to under-densified representations and suboptimal reconstruction performance.

To address these challenges, we propose PhysHDR-GS, a physically inspired HDR-NVS framework that models scene appearance as intrinsic reflectance and adjustable ambient illumination.  Motivated by different response patterns in \cref{fig:teaser}, our framework introduces two complementary branches: an image-exposure (IE) branch that modulates exposure on captured images, and a Gaussian-illumination (GI) branch that modulates the ambient illumination of 3D Gaussians. This dual modulation preserves fidelity to standard camera observations while explicitly capturing ambient illumination-dependent changes in appearance. During training, to provide explicit supervision for HDR content, we impose a cross-branch HDR consistency loss between  HDR outputs of IE and GI branch, and further leverage a learnable tone mapper to fuse tonemapped results as LDR outputs. Moreover, to mitigate the exposure-biased gradient starvation, we introduce an illumination-guided gradient scaling strategy that amplifies per-Gaussian gradients based on illumination deviation, preventing under/over-exposed Gaussians from being under-densified.  Our contributions are summarized as follows:
\begin{itemize}
\item We propose a physically inspired HDR-NVS framework that combines an image-exposure (IE) branch and a Gaussian-illumination (GI) branch to reconstruct standard camera observations and explicitly capture illumination-dependent appearance changes, respectively.

\item An HDR consistency loss is imposed between IE and GI branches to enable explicit HDR supervision without ground truth. A cross-fusion-based tone mapper further fuses LDR results to improve reconstruction quality.

\item We propose an illumination-guided gradient scaling strategy that amplifies Gaussian gradients based on illumination deviation, alleviating gradient starvation and reducing under-densified representations.

\item Experimental results on two exposure settings across three benchmarks demonstrate our superiority in reconstructing HDR details, while maintaining real-time rendering speed (up to \textbf{76 FPS}). Ablation studies further demonstrate the effectiveness of each component.   

\end{itemize}

\vspace{-4pt}\section{Related Work}\label{sec:related} \vspace{-2pt}
\subsection{High Dynamic Range Novel View Synthesis}\vspace{-2pt}
Early NeRF-based HDR-NVS methods~\cite{huang2022hdr,jun2022hdr,huang2024ltm,wang2024cinematic} reconstruct an HDR radiance field from multi-exposure LDR inputs, with some  exploring RAW space~\cite{mildenhall2022nerf} to better preserve dynamic range. However, volumetric rendering makes both training and inference time-consuming. Recent 3DGS-based methods~\cite {singh2024hdrsplat, jin2024lighting,li2024chaos,li2024chaos} model scenes with Gaussian primitives and adopt rasterization-based rendering, achieving significant acceleration. HDR-GS~\cite{cai2024hdr} fits HDR color with spherical harmonics and predicts exposure-conditioned LDR views via an MLP-based tone mapper.  Wu~\etal~\cite{wu2024hdrgs} include luminance dimension for irradiance-to-color conversion and employ an asymmetric grid for tone mapping.  To stabilize 3D tone mapping, GaussHDR~\cite{liu2025gausshdr} unifies 3D and 2D local tone mapping and fuses dual-branch LDR outputs. Currently, HDR-NVS has also been extended to multi-modal settings~\cite{li2024learning,chen2025evhdr},  single-exposure~\cite{li2025sehdr,zhang2025high} and inconsistent illumination scenarios~\cite{cui2025luminance, gong2025casualhdrsplat,bolduc2025gaslight}. However, most existing methods still follow a conventional HDR imaging pipeline, where different lighting levels are simulated by applying exposure and tone mapping to 2D images.  Without modeling illumination in 3D space,  environment-dependent attributes of the scene are largely underexplored. Therefore, we explicitly model lighting-conditioned scene appearance by jointly controlling camera exposure and ambient illumination, revealing complementary HDR details in image space and 3D radiance.

\subsection{High Dynamic Range Reconstruction}
HDRCNN~\cite{eilertsen2017hdr} makes an early exploration on reconstructing HDR from a single LDR image with a CNN.  
HDRNet~\cite{gharbi2017deep} achieves real-time enhancement by learning a low-resolution-based affine color transform and applying it to the full-resolution image. For multi-exposure HDR imaging, recent methods~\cite{chen2023improving,li2025afunet,debevec2023recovering} typically decompose this problem into alignment, fusion, and reconstruction. AFUNet~\cite{li2025afunet} introduces a cross-iterative network that alternates alignment and fusion to progressively reconcile motion and exposure discrepancies. Debevec~\etal~\cite{debevec2023recovering} estimate the camera response curve from bracketed exposures under reciprocity and then fuse multiple photographs.  Le~\etal~\cite{Le_2023_WACV} invert the camera response and synthesize multi-exposure images to hallucinate missing details. SAFNet~\cite{kong2024safnet} selectively estimates cross-exposure motion and valuable region masks with shared decoders and performs explicit fusion. Generative approaches~\cite{zeng2020sr,niu2021hdr,wang2023glowgan,wang2025lediff} further leverage powerful priors to recover details in saturated or underexposed regions. Despite impressive HDR reconstruction demonstrated by these methods, they are essentially 2D methods and are unable to understand 3D scene, thus cannot synthesize novel HDR views.

\begin{figure*}[t]
    \centering
    \includegraphics[width=1\linewidth]{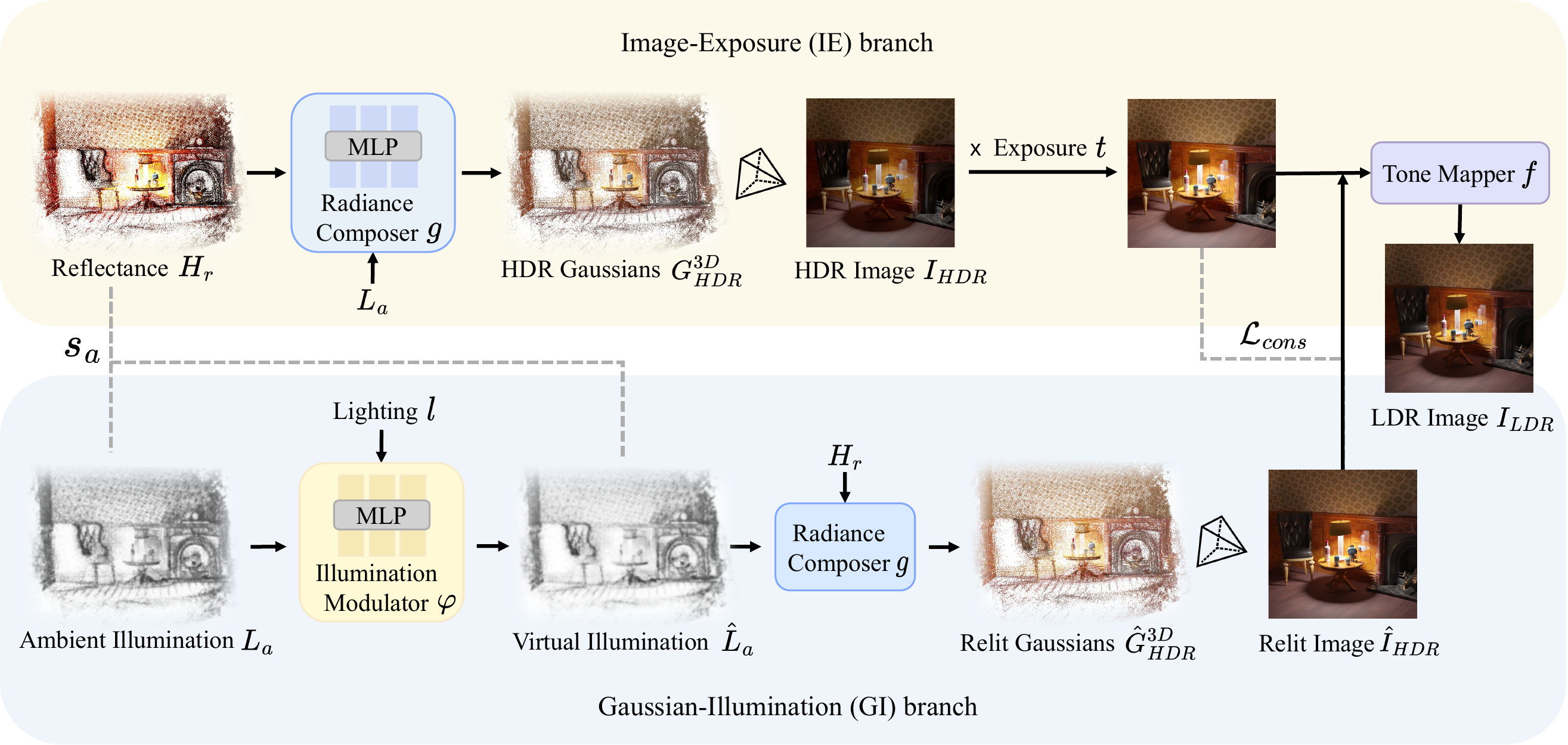}\vspace{-2pt}
    \caption{Overview of the proposed PhysHDR-GS, where Gaussian color is modeled from intrinsic reflectance and ambient illumination. The image–exposure (IE) branch modulates exposure $t$ on 2D images, while the Gaussian–illumination (GI) branch modulates ambient illumination $L_a$ on 3D Gaussians, yielding complementary dynamic-range details. Tone mapper $f$ performs tone mapping and dual-branch fusion for final LDR results. During training, a cross-branch HDR consistency loss $\mathcal{L}_{cons}$ enables explicit HDR self-supervision. Illumination-Guided Gradient Scaling rescales per-Gaussian gradients with $s_a$ to mitigate under-splitting in extreme exposure regions.}\vspace{-6pt}
    \label{fig:framework}
\end{figure*}

\vspace{-4pt}\section{Preliminaries}\label{sec:pre}\vspace{-2pt}
\subsection{3D Gaussian Splatting}\vspace{-2pt}
3DGS~\cite{3dgs} model a static 3D scene with a set of explicit Gaussian primitives $G^{3D}$ as follows:
\vspace{-4pt} \begin{equation} \label{eq:3dgs}
G^{3D} = \{(\boldsymbol{\mu}_i,\boldsymbol{\Sigma}_i,\boldsymbol{\alpha}_i,\boldsymbol{c}_i)\}_{i=1}^{N},
\end{equation} 
where N is the number of Gaussians. $\boldsymbol{\mu}_i$, $\boldsymbol{\Sigma}_i$, $\boldsymbol{\alpha}_i$, $\boldsymbol{c}_i$ denote the center position, covariance, opacity and color of $i$-th Gaussian primitive, respectively. $\boldsymbol{\Sigma}_i$ is defined with a scaling matrix $\boldsymbol{S}_i$ and rotation matrix $\boldsymbol{R}_i$ as $\boldsymbol{R}_i\boldsymbol{S}_i\boldsymbol{S}_i^\top\boldsymbol{R}_i^\top$.  During rendering, given the projection matrix $\boldsymbol{P}$, extrinsic matrix $\boldsymbol{W}$ and Jacobian of projective transformation $\boldsymbol{J}$,  each 3D Gaussian primitive in world space is projected as a 2D Gaussian $G^{2D}$, with transformed covariance $\hat{\boldsymbol{\Sigma}}=\boldsymbol{J} \boldsymbol{W} \boldsymbol{\Sigma} \boldsymbol{W}^\top \boldsymbol{J}^\top$ and camera coordinate  $\hat{\boldsymbol{\mu}}=\boldsymbol{P}\boldsymbol{W}[\boldsymbol{\mu}, 1]^{\top} \in\mathbb{R}^4$. Then the 2D pixel $\mathbf{p}$ is shaded as follows:
\begin{equation}   \label{eq:densify}
\resizebox{0.43\textwidth}{!}{$
\boldsymbol{C}(\mathbf{p})=\sum_{i = 1}^{N}  \boldsymbol{\alpha}_i \boldsymbol{c}_i G_i^{2D}(\mathbf{p}) \prod_{j=1}^{i-1}\left(1-\boldsymbol{\alpha}_j G_j^{2D}(\mathbf{p}) \right).
$}
\end{equation} During training, the $i$-th primitive is densified if  average screen-space gradient exceeds a threshold $\tau_p$:
\begin{equation}\label{eq:densify}
\frac{1}{M_i}\sum_{k=1}^{M_i}
\left\|
\frac{\partial \mathcal{L}_k}{\partial \boldsymbol{\mu}_{i,k}^{\text{ndc}}}
\right\|_2
>\tau_p,
\end{equation}
where $M_i$ is the number of visible views, and $\mathcal{L}_k$ denote per-view loss. $\boldsymbol{\mu}^{\text{ndc}}
= \frac{\hat{\boldsymbol{\mu}}_{1:3}}{\hat{\mu}_4}$ is the normalized device coordinates (NDC) center.

\subsection{Physically-Based Rendering}
Physically based rendering (PBR) models object color as the interaction between illumination and surface material. For a surface point $\mathbf{x}$ with normal $\mathbf{n}$, outgoing radiance toward direction $\boldsymbol{\omega}_o$ is given by rendering equation~\cite{kajiya1986rendering}:
\begin{equation}\label{eq:rendering}
\resizebox{0.43\textwidth}{!}{$
L_o\left(\mathbf{x}, \boldsymbol{\omega}_o\right)=L_e\left(\mathbf{x}, \boldsymbol{\omega}_o\right)+\int_{\Omega} f_r\left(\mathbf{x}, \boldsymbol{\omega}_i, \boldsymbol{\omega}_o\right) L_i\left(\mathbf{x}, \boldsymbol{\omega}_i\right)\left(\mathbf{n}\boldsymbol{\omega}_i\right) d \boldsymbol{\omega}_i,
$}
\end{equation}
where $L_e$ is emitted radiance, $L_i$ is incident radiance from direction $\boldsymbol{\omega}_i$, $f_r$ is the bidirectional reflectance distribution function (BRDF). $\Omega$ is the upper hemisphere around $\mathbf{n}$. With uniform hemispherical illumination (\ie, $L_i(\mathbf{x},\boldsymbol{\omega}_i)\equiv L_a(\mathbf{x})$ for all $\boldsymbol{\omega}_i\in\Omega$) and direction-independent emission (\ie, $L_e(\mathbf{x},\boldsymbol{\omega}_o)\equiv L_e(\mathbf{x})$),  \cref{eq:rendering} simplifies to:
\begin{equation}\label{eq:gen-dh}
\begin{aligned}
L_o(\mathbf{x},\boldsymbol{\omega}_o) &= L_e(\mathbf{x}) + L_a(\mathbf{x})\,H_r(\mathbf{x},\boldsymbol{\omega}_o),\\
\end{aligned}
\end{equation}
where $H_r(\mathbf{x},\boldsymbol{\omega}_o) = \int_{\Omega} f_r(\mathbf{x},\boldsymbol{\omega}_i,\boldsymbol{\omega}_o)\,(\mathbf{n}\boldsymbol{\omega}_i)\,d\boldsymbol{\omega}_i$ is the hemispherical-directional reflectance.  A LDR pixel at location \(\mathbf{p}\) is captured by applying exposure $t$ and camera response function (CRF) $f$ to the accumulated radiance:
\begin{equation}\label{eq:ldr-generalv1}
\resizebox{0.43\textwidth}{!}{$
I_{LDR}(\mathbf{p})=
f\!\big(t\,L_o(\mathbf{x},\boldsymbol{\omega}_o)\big) =
f\!\big(t\,L_e(\mathbf{x}) + t\,L_a(\mathbf{x})\,H_r(\mathbf{x},\boldsymbol{\omega}_o)\big),
$}
\end{equation} where $L_e(\mathbf{x})$ and $H_r(\mathbf{x},\boldsymbol{\omega}_o)$ are scene-intrinsic and exposure-invariant, while $t$ and $L_a(\mathbf{x})$ scales signal before CRF $f$.  We model $L_o(\mathbf{x},\boldsymbol{\omega}_o)$ with learnable parameters $g$, LDR pixel $\mathbf{p}$ thus depends on \(t\) and \(L_a\) as follows:
\begin{equation}\label{eq:ldr-general}
I_{LDR}(\mathbf{p};t,L_a) = f\!\big( t \cdot g\left(L_a(\mathbf{x}), H_r(\mathbf{x},\boldsymbol{\omega}_o)\right) \big),
\end{equation}
where constant $L_e$ is absorbed. \cref{eq:ldr-general} shows that exposure $t$ and ambient illumination $L_a$ play complementary roles in shaping the dynamic range of the pre‑CRF signal.

\section{Method}\label{sec:method}
Given a set of multi-exposure LDR views, we aim to reconstruct a 3D scene with HDR details. As shown in \cref{fig:framework}, Gaussian color is factorized into intrinsic reflectance $H_r$ and adjustable ambient illumination $L_a$.  To capture complementary dynamic details from exposure and ambient illumination, we introduce an image-exposure (IE) branch that follows the camera pipeline to modulate exposure $t$ on 2D images, and a Gaussian–illumination (GI) branch that modulates ambient illumination $L_a$ on 3D Gaussians. We elaborate the method from three aspects: physical radiance composition, self‑consistent HDR fusion, and illumination‑guided gradient scaling.

\subsection{Physical Radiance Composition}
\noindent \textbf{Image-exposure (IE) branch.}
As shown in \cref{fig:framework}, given separately modeled reflectance $H_r$ and ambient illumination $L_a$, the Gaussian color $\boldsymbol{c}$ is produced by an MLP-based radiance composer $g$: 
\vspace{-2pt}\begin{equation}\label{eq:compose}
\boldsymbol{c}=g(L_a,H_r).\vspace{-2pt}
\end{equation}
Collecting geometry ($\boldsymbol{\mu}, \boldsymbol{\Sigma}$), opacity $\boldsymbol{\alpha}$ and color $\boldsymbol{c}$ further yields the HDR Gaussians set:
\vspace{-4pt}\begin{equation} \label{eq:3dgs_hdr} 
G^{3D}_{HDR} = \{(\boldsymbol{\mu}_i,\boldsymbol{\Sigma}_i,\boldsymbol{\alpha}_i,\boldsymbol{c}_i)\}_{i=1}^{N}.
\end{equation}
Given a target view, the HDR Gaussian set $G^{3D}_{HDR}$ is projected to HDR image $I_{HDR}$. To cover different luminance bands and bring mid-tone regions into the camera’s responsive range, the IE branch applies exposure $t$ on $I_{HDR}$ (\ie, $I_{HDR}\times t$) to globally scale the pre-tone-mapping signal.

\noindent \textbf{Gaussian-illumination (GI) branch.}
With the disentanglement of reflectance and illumination, the 3D scene can be relit by modulating $L_a$ based a target lighting condition. We introduce an illumination modulator $\varphi$ to produce virtual illumination $\hat{L}_a$ as follows:
\vspace{-4pt}\begin{equation}
\hat{L}_a=\varphi(L_a, l),
\end{equation}
where $l$ denotes the target lighting level. Replacing $L_a$ with $\hat{L}_a$ in \cref{eq:compose} gives relit color $\hat{\boldsymbol{c}}$ and the relit Gaussians:
\vspace{-2pt}\begin{equation} \label{eq:3dgs_hdr}
\hat{G}^{3D}_{HDR} = \{(\boldsymbol{\mu}_i,\boldsymbol{\Sigma}_i,\boldsymbol{\alpha}_i,\hat{\boldsymbol{c}}_i)\}_{i=1}^{N},\vspace{-2pt}
\end{equation}
which is further projected to relight image $\hat{I}_{HDR}$ with the same viewpoint of $I_{HDR}$. By adjusting virtual illumination $\hat{L}_a$, the GI branch enables rescaling radiance intensity locally to avoid saturation. Together, IE and GI branches provide better coverage of a higher dynamic range.

\begin{figure}[t]
    \centering
   \vspace{-4pt} \includegraphics[width=1\linewidth]{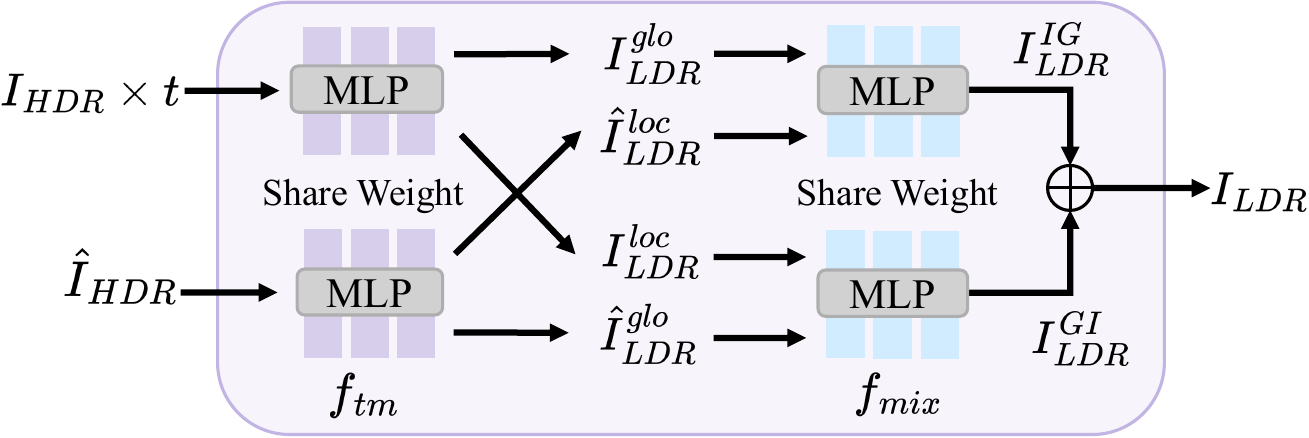}\vspace{-2pt}
    \caption{Illustration of the tone mapper $f$. Given inputs $I_{HDR}\times t$ and $\hat{I}_{HDR}$, the tone-mapping MLP $f_{tm}$ first predicts global and local LDR outputs. The fusion MLP $f_{mix}$ then cross-fuses these global-local pairs to produce the final LDR result $I_{LDR}$.}
    \label{fig:tonemapper}
\end{figure}

\subsection{Self-Consistent HDR Fusion}\label{sec:self super}
\noindent \textbf{Tone-mapped LDR learning.}
As illustrated in \cref{fig:framework}, imposing global exposure  $t$ and local virtual ambient illumination $\hat{L}_a$ yields HDR signals $I_{HDR}\times t$ and $\hat{I}_{HDR}$. To aggregate complementary dynamic range details and supervise with standard camera-captured LDR images, both  HDR signals are further mapped into the LDR domain. As illustrated in \cref{fig:tonemapper}, the proposed tone mapper $f$ consists of two lightweight MLPs $f_{tm}$ and $f_{mix}$, which perform global-local tone mapping and LDR fusion, respectively. Given an HDR image, $f_{tm}$ conducts tone mapping and outputs a pair of global-local LDR images. Applying $f_{tm}$ to the exposure-scaled $I_{HDR}\times t$  and relit $\hat{I}_{HDR}$ thus yields intermediate LDR predictions \{$I^{glo}_{LDR}$, $I^{loc}_{LDR}$, $\hat{I}^{glo}_{LDR}$, $\hat{I}^{loc}_{LDR}$\}. The fusion MLP $f_{mix}$ then conducts cross-fusion: 
\vspace{-2pt}\begin{equation}
     I^{IG}_{LDR}=f_{mix}(I^{glo}_{LDR},\hat{I}^{loc}_{LDR}),\vspace{-2pt}
\end{equation}
\vspace{-4pt}\begin{equation}
    I^{GI}_{LDR}=f_{mix}(I^{glo}_{LDR},I^{loc}_{LDR}).
\end{equation}
The final LDR prediction $I_{LDR}$ is obtained by aggregating the fused outputs:
\vspace{-6pt}\begin{equation}
    I_{LDR} =I^{IG}_{LDR}+I^{GI}_{LDR}.\vspace{-2pt}
\end{equation}
To supervise the results with available LDR views, we define the reconstruction loss as follows:
\vspace{-2pt}
\begin{equation}
\resizebox{0.42\textwidth}{!}{$
 \mathcal{L}_{rec}
    =
    \sum_{I \in \mathcal{I}}
    \Big[
        \gamma\,\mathcal{L}_{MSE}(I, I_{gt})
        +
        \mathcal{L}_{D\text{-}SSIM}(I, I_{gt})
    \Big],
    $}\vspace{-2pt}
\end{equation}
where \(\mathcal{I} = \{I_{LDR},\, I^{IG}_{LDR},\, I^{GI}_{LDR}\}\), \(I_{gt}\) is LDR ground truth. Hyperparameter \(\gamma\) controls the weight of MSE term.

\noindent \textbf{Self-consistent HDR learning.}
Tone-mapping HDR images into LDR inevitably compresses dynamic range and clips extreme values (\eg, saturated highlights and deep shadows). As a result, abnormal HDR values may not be effectively constrained by LDR supervision. To apply explicit supervision to HDR, we impose a cross-branch HDR consistency loss between the IE and GI branch.  As shown in \cref{fig:framework}, for each view, we compute the consistent loss with exposure-scaled $I_{HDR} \times t$ and relit image $\hat{I}_{HDR}$ as follows,
\vspace{-4pt}\begin{equation}
    \mathcal{L}_{\mathrm{cons}}
    =
    \big\|
        \mathcal{G}\big(I_{HDR} \times t\big)
        -
        \mathcal{G}\big(\hat{I}_{HDR}\big)
    \big\|_1,
\end{equation}
where $\mathcal{G}$ indicates applying Gaussian blur to avoid penalizing misaligned details. In practice, we set lighting level $l=t$ so that $I_{HDR} \times t$ and $\hat{I}_{HDR}$ are comparable in brightness. 
$\mathcal{L}_{\mathrm{cons}}$ therefore matches the overall illumination and low-frequency structure of the HDR predictions, providing explicit supervision for the HDR content.

\begin{figure}[t]
    % \centering
  \hspace{-6pt} \includegraphics[width=1.02\linewidth, clip, trim=16pt 18pt 8pt 6pt]{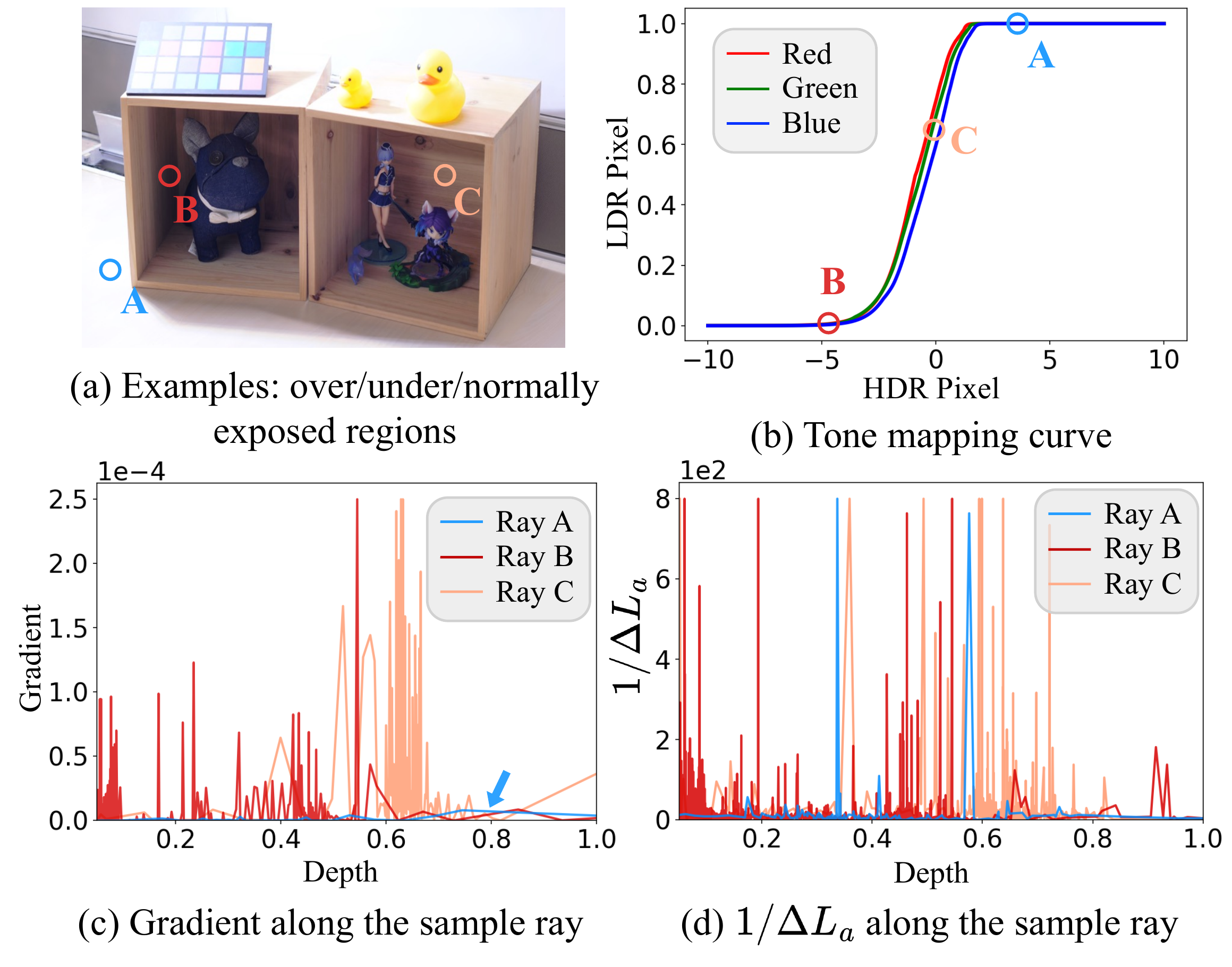}\vspace{-2pt}
    \caption{Gradient and illumination deviation analysis, where over/under-exposed pixels lie in flat regions of the tone mapping curve and yield a small Gaussian gradient. The gradient shows positive correlation with reciprocal illumination deviation $1/\Delta L_a$.}
    \label{fig:gradient_scaling}
\end{figure}

\subsection{Illumination-Guided Gradient Scaling}\label{sec:gradient}
Given multi-exposure views of the same scene, pixels can be categorized into three types: over-exposed, under-exposed and normally-exposed regions (\eg, point A, B, C in \cref{fig:gradient_scaling}(a)). The tone mapping curve (shown in \cref{fig:gradient_scaling}(b)) typically compresses dynamic range, clipping highlights and shadows. Consequently, compared with normally-exposed regions (\eg, point C), over-exposed  (\eg, point B) and under-exposed regions (\eg, point A) lie in flat region of CRF (see \cref{fig:gradient_scaling}(b)) and yiled much smaller slopes.  As shown in \cref{fig:gradient_scaling}(c), Gaussian primitives that model these regions receive very limited gradient. Due to the gradient-based densification shown in \cref{eq:densify}, these Gaussians seldom reach the threshold $\tau_p$, resulting in insufficient splitting and under-densified representations~\cite{zhang2024pixel,philip2023floaters,zeng2026arbitrary}.

We observe that the illumination deviation of a Gaussian primitive is inversely correlated to the received gradient. As shown in \cref{fig:gradient_scaling}(d),  we visualize the reciprocal of illumination deviation $1/\Delta L_a$, where $\Delta L_a=\lvert L_a-\hat{L}_a\lvert$ measures the illumination discrepancy. Compared with normally exposed regions, over-exposed/under-exposed regions exhibit much larger illumination deviation (\ie, smaller $1/\Delta L_a$), which is negatively related to their gradient magnitudes. This suggests an illumination deviation-based strategy to compensate for gradient deficiency. Based on this observation, we propose the illumination-guided gradient scaling, which adaptively rescales per-Gaussian gradients.  The gradient scaling factor $s_a$ is defined as:
\begin{equation}
    s_a \;=\; s \cdot \sigma\!\big(\lvert L_a - \hat{L}_a \rvert\big) + 1,
\end{equation}
where \(\sigma(\cdot)\) is the sigmoid function, \(s\) is a hyperparameter controlling the maximum scaling strength. Therefore, the densification criterion in \cref{eq:densify} is revised below:
\begin{equation}\label{eq:densify-scale}
   \mathbb{I}_{i}\left(s_a \right)\frac{1}{M_i}\sum_{k=1}^{M_i} 
\left\|
\frac{\partial \mathcal{L}_k}{\partial \boldsymbol{\mu}_{i,k}^{\text{ndc}}}
\right\|_2
>\tau_{p},
\end{equation}
where $\mathbb{I}_{i}\left(s_a \right)$ indicates retrieving the scaling factor for $i$-th Gaussian primitive. By amplifying gradients according to the illumination deviation,  illumination-guided gradient scaling effectively prevents Gaussians in over/under-exposed regions from insufficient splitting, leading to superior reconstruction performance.

\subsection{Loss Function}
For each training view, the total loss combines the LDR reconstruction loss and the HDR consistency loss. For synthetic datasets, we  follow~\cite{cai2024hdr,huang2022hdr,liu2025gausshdr} to additionally impose a uniform-exposure regularization:
\begin{equation}
    \mathcal{L}_{\text{total}}
    =
    \lambda_1\,\mathcal{L}_{\text{rec}}
    +
    \lambda_2\mathcal{L}_{\text{cons}}
    +
    \lambda_3\mathcal{L}_{\text{unit}},
\end{equation}
where $\lambda_1$, $\lambda_2$  and $\lambda_3$ are weight of each loss term.

\begin{table*}[t]
\centering
  \caption{Quantitative results on realistic HDR-NeRF-Real~\cite{huang2022hdr} and HDR-Plenoxels-Real~\cite{jun2022hdr} datasets, where the best and second-best results are highlighted in \colorbox{red!30}{red} and \colorbox{yellow!30}{yellow}, respectively. $\dagger$ indicates variants built on Scaffold-GS~\cite{lu2024scaffold}. The proposed method achieves the overall best performance, demonstrating its effectiveness in synthesizing high-quality novel views across different exposure levels.} 
\label{tab:performance_real}
\Large
\resizebox{1\linewidth}{!}{
\begin{tabular}{c|l|ccc ccc|ccc ccc}
\toprule
 \multicolumn{2}{c|}{\multirow{3}{*}{Method}} & \multicolumn{6}{c|}{HDR-NeRF-Real} & \multicolumn{6}{c}{HDR-Plenoxels-Real}\\
\cmidrule(lr){3-8} \cmidrule(lr){9-14} 
  \multicolumn{2}{c|}{}  & \multicolumn{3}{c}{LDR-OE ( $t_1, t_3, t_5$ )} & \multicolumn{3}{c|}{LDR-NE ( $t_2, t_4$ )} & \multicolumn{3}{c}{LDR-OE ( $t_1, t_3, t_5$ )} & \multicolumn{3}{c}{LDR-NE ( $t_2, t_4$ )} \\
\cmidrule(lr){3-5} \cmidrule(lr){6-8} \cmidrule(lr){9-11} \cmidrule(lr){12-14}
  \multicolumn{2}{c|}{} & PSNR $\uparrow$ & SSIM $\uparrow$ & LPIPS $\downarrow$ & PSNR $\uparrow$ & SSIM $\uparrow$ & LPIPS $\downarrow$ & PSNR $\uparrow$ & SSIM $\uparrow$ & LPIPS $\downarrow$ & PSNR $\uparrow$ & SSIM $\uparrow$ & LPIPS $\downarrow$ \\
\midrule 
 \multirow{6}{*}{\rotatebox{90}{exp3}} & HDR-NeRF~\cite{huang2022hdr}  & 34.27 & 0.9532 & 0.063 & 32.15 & 0.9475 & 0.074& - & - & - & - & - & - \\
& HDR-GS~\cite{cai2024hdr} &  34.87 & 0.9697 & 0.021 & 31.02 & 0.9636 & 0.029 &31.17 & 0.9509 & 0.040 & 28.60 & 0.9285 & 0.051\\
& GaussHDR~\cite{liu2025gausshdr} & 36.05 & 0.9739 & 0.015 & 33.49 & 0.9707 & 0.017 & 31.50 & 0.9530 & 0.037 & 28.92 & 0.9326 & 0.044\\
& GaussHDR$\dagger$~\cite{liu2025gausshdr} &  \cellcolor{yellow!30}36.32 & 0.9770 & 0.011 & \cellcolor{yellow!30}33.84 & \cellcolor{yellow!30}0.9738 & 0.014 & \cellcolor{yellow!30}32.87 & \cellcolor{yellow!30}0.9590 & \cellcolor{yellow!30}0.028 & \cellcolor{red!30}29.78 & \cellcolor{red!30}0.9391 & \cellcolor{yellow!30}0.037\\
& Ours & 36.19 & \cellcolor{red!30}0.9779 & \cellcolor{yellow!30}0.011 & 33.68 & \cellcolor{red!30}0.9738 & \cellcolor{yellow!30}0.014 & 32.19 & 0.9536 & 0.033 & 28.79 & 0.9299 & 0.046\\
& Ours$\dagger$& \cellcolor{red!30}36.91 & \cellcolor{yellow!30}0.9777 & \cellcolor{red!30}0.009 & \cellcolor{red!30}34.15 & 0.9737 & \cellcolor{red!30}0.012  &  \cellcolor{red!30}33.06 & \cellcolor{red!30}0.9592 & \cellcolor{red!30}0.025 & \cellcolor{yellow!30}29.76 & \cellcolor{yellow!30}0.9389 & \cellcolor{red!30}0.034\\ 
\midrule
 \multirow{6}{*}{\rotatebox{90}{exp1}} & HDR-NeRF~\cite{huang2022hdr}   &34.26 & 0.9532 & 0.063 & 31.55 & 0.9483 & 0.074  & - & - & - & - & - & -  \\
& HDR-GS~\cite{cai2024hdr} & 32.96 & 0.9597 & 0.028 & 29.66 & 0.9535 & 0.035 & 29.62 & 0.9373 & 0.051 & 27.32 & 0.9151 & 0.060\\
& GaussHDR~\cite{liu2025gausshdr} & 34.59 & 0.9678 & 0.018 & 32.98 & 0.9653 & 0.020 & 30.87 & 0.9458 & 0.041 & 28.26 & 0.9230 & 0.050\\
& GaussHDR$\dagger$~\cite{liu2025gausshdr}& 34.63 & \cellcolor{red!30}0.9710 & \cellcolor{yellow!30}0.014 & \cellcolor{yellow!30}33.29 & \cellcolor{red!30}0.9684  & \cellcolor{yellow!30}0.016  & \cellcolor{yellow!30}32.24 & \cellcolor{yellow!30}0.9540 & \cellcolor{yellow!30}0.031 & \cellcolor{yellow!30}28.88 & \cellcolor{red!30}0.9315 & \cellcolor{yellow!30}0.041\\
& Ours & \cellcolor{yellow!30}34.64 & \cellcolor{yellow!30}0.9707 & 0.015 & 32.64 & 0.9672 & 0.018 & 31.20 & 0.9464 & 0.037 & 27.59 & 0.9193 & 0.047\\
& Ours$\dagger$& \cellcolor{red!30}34.84 & 0.9705 & \cellcolor{red!30}0.012 & \cellcolor{red!30}33.40 & \cellcolor{yellow!30}0.9677 & \cellcolor{red!30}0.014 & \cellcolor{red!30}32.34 & \cellcolor{red!30}0.9540 & \cellcolor{red!30}0.029 & \cellcolor{red!30}29.06 & \cellcolor{yellow!30}0.9313 & \cellcolor{red!30}0.038\\  
\bottomrule
\end{tabular}
} 
\end{table*}

\begin{table*}[t]
\centering
  \caption{Quantitative results on the synthetic HDR-NeRF-Syn~\cite{huang2022hdr} dataset, where the best and second-best results are highlighted in \colorbox{red!30}{red} and \colorbox{yellow!30}{yellow}, respectively. $\dagger$ indicates variants built on Scaffold-GS~\cite{lu2024scaffold}. Our method consistently outperforms baselines on both LDR and HDR scenarios, demonstrating its effectiveness in reconstructing HDR details and preserving information during tonemapping.}
\label{tab:performance_syn}
% \small
\resizebox{0.96\linewidth}{!}{
\begin{tabular}{c|l|ccc ccc  ccc}
\toprule
 \multicolumn{2}{c|}{\multirow{2}{*}{Method}}  & \multicolumn{3}{c}{LDR-OE ( $t_1, t_3, t_5$ )} & \multicolumn{3}{c}{LDR-NE ( $t_2, t_4$ )} & \multicolumn{3}{c}{HDR}  \\
\cmidrule(lr){3-5}\cmidrule(lr){6-8} \cmidrule(lr){9-11}
  \multicolumn{2}{c|}{} & PSNR $\uparrow$ & SSIM $\uparrow$ & LPIPS $\downarrow$ & PSNR $\uparrow$ & SSIM $\uparrow$ & LPIPS $\downarrow$ & PSNR $\uparrow$ & SSIM $\uparrow$ & LPIPS $\downarrow$  \\
 \midrule  
 \multirow{6}{*}{\rotatebox{90}{exp3}} & HDR-NeRF~\cite{huang2022hdr}   & 38.82 & 0.9657 & 0.032 & 38.07 & 0.9641 & 0.034 & 26.63 & 0.9523 & 0.046\\
& HDR-GS~\cite{cai2024hdr} & 40.28 & 0.9781 & 0.018 & 27.07 & 0.8744 & 0.127 & 17.51 & 0.6982 & 0.205\\
& GaussHDR ~\cite{liu2025gausshdr}& 42.28 & 0.9853 & 0.007 & 41.65 & 0.9850 & 0.007 & 37.78 & 0.9704 & 0.017\\
& GaussHDR$\dagger$~\cite{liu2025gausshdr} & \cellcolor{yellow!30}43.87 & \cellcolor{yellow!30}0.9899 & \cellcolor{yellow!30}0.004 & \cellcolor{yellow!30}42.74 & \cellcolor{yellow!30}0.9894 & \cellcolor{yellow!30}0.004 & \cellcolor{yellow!30}39.08 & \cellcolor{yellow!30}0.9767 &\cellcolor{yellow!30} 0.011\\
& Ours & 43.11 & 0.9848 & 0.008 & 42.48 & 0.9852 & 0.008 & 38.77 & 0.9710 & 0.016\\
& Ours$\dagger$& \cellcolor{red!30}44.26 & \cellcolor{red!30}0.9899 & \cellcolor{red!30}0.003 & \cellcolor{red!30}43.19 & \cellcolor{red!30}0.9896 & \cellcolor{red!30}0.004 & \cellcolor{red!30}39.21 & \cellcolor{red!30}0.9768 & \cellcolor{red!30}0.010\\ 
\toprule
 \multirow{6}{*}{\rotatebox{90}{exp1}} & HDR-NeRF~\cite{huang2022hdr}   & 38.68 & 0.9649 & 0.032 & 37.63 & 0.9621 & 0.035 & 26.61 & 0.9523 & 0.046\\
& HDR-GS~\cite{cai2024hdr} & 38.22 & 0.9688 & 0.026 & 25.42 & 0.8639& 0.136 & 16.43 & 0.6048 & 0.265\\
& GaussHDR~\cite{liu2025gausshdr} & 41.67 & 0.9839 & 0.007 & 41.21 & 0.9840 & 0.008 & 37.44 & 0.9691 & 0.018 \\
& GaussHDR$\dagger$~\cite{liu2025gausshdr} &  \cellcolor{red!30}42.94 & \cellcolor{red!30}0.9883 & \cellcolor{yellow!30}0.004 & \cellcolor{yellow!30}42.01 & \cellcolor{yellow!30}0.9879 & \cellcolor{yellow!30}0.005 & \cellcolor{yellow!30}38.62 & \cellcolor{yellow!30}0.9752 & \cellcolor{yellow!30}0.011\\
& Ours & 41.99 & 0.9835 & 0.006 & 41.47 & 0.9842 & 0.007 & 38.34 & 0.9701 & 0.013\\
& Ours$\dagger$& \cellcolor{yellow!30}42.80 & \cellcolor{yellow!30}0.9881 & \cellcolor{red!30}0.004 & \cellcolor{red!30}42.13 & \cellcolor{red!30}0.9882 & \cellcolor{red!30}0.004 & \cellcolor{red!30}38.79 & \cellcolor{red!30}0.9756 & \cellcolor{red!30}0.010\\ 
\bottomrule
\end{tabular}
} 
\end{table*}

\section{Experiment}\label{sec:exp}
\subsection{Experimental Settings}
\noindent \textbf{Datasets and settings.}
Experiments are conducted on three benchmarks: realistic HDR-NeRF-Real~\cite{huang2022hdr}, HDR-Plenoxels-Real~\cite{jun2022hdr} and synthesized HDR-NeRF-Syn~\cite{huang2022hdr}. All datasets provide LDR views at five exposure times $\{t_1,t_2,t_3,t_4,t_5\}$, which are grouped into LDR-OE ($\{t_1,t_3,t_5\}$) and LDR-NE ($\{t_2,t_4\}$). HDR-NeRF-Syn additionally provides HDR ground truth for evaluating HDR content. Note that these HDR ground-truth images are only used for evaluation and are not used for training by any method. For HDR-NeRF-Real and HDR-NeRF-Syn dataset, we train with 18 LDR-OE views and evalute with remaining 17 views. For HDR-Plenoxels-Real dataset,  we train with 27 LDR-OE views and evaluate with remaining 13 views. We follow GaussHDR~\cite{liu2025gausshdr} to train with two exposure settings: randomly assigning exposure at $\{t_1,t_3,t_5\}$  at every iteration, (denoted as exp3), and initializing with random exposure from $\{t_1,t_3,t_5\}$ and keeping fixed throughout training (denoted as exp1).  

\noindent \textbf{Baselines and model variants.}
We compared against both NeRF-based (\ie, HDR-NeRF~\cite{huang2022hdr}) and 3DGS-based methods. For the 3DGS-based methods methods, we include HDR-GS~\cite{cai2024hdr} and GaussHDR~\cite{liu2025gausshdr}).To assess the generality, we follow GaussHDR to integrate proposed components into both vanilla 3DGS backbone and Scaffold-GS~\cite{lu2024scaffold}. Variants built on Scaffold-GS are marked with $\dagger$.

\noindent \textbf{Metrics.} We evaluate reconstruction quality using PSNR and SSIM for pixel-wise fidelity, and LPIPS~\cite{zhang2018unreasonable} for perceptual quality. HDR views are first tone-mapped with $\mu$-law and then conduct quantitative evaluation~\cite{liu2025gausshdr,huang2022hdr}.

\noindent \textbf{Implementation details.}
We train each scene for 30k iterations. During the first 10k iterations, we freeze the fusion MLP $f_{mix}$ and only train the tone-mapping MLP $f_{tm}$. $f_{mix}$ is then unfrozen for the remaining iterations.  The weights $\lambda_1$, $\lambda_2$ and $\lambda_3$ are set to 1, 0.5 and 0, respectively. For synthetic scenes, we additionally enable the unit-exposure regularization term by setting $\lambda_3$ set to 0.5. $\gamma$ is set to 0.2. Scaling hyperparameter $s$ is set to 1. All experiments are conducted with a single A6000 GPU. More details are included in the supplementary materials.

\begin{figure*}
    \centering
        \includegraphics[width=1\linewidth,clip, trim=20pt 30pt 20pt 18pt]{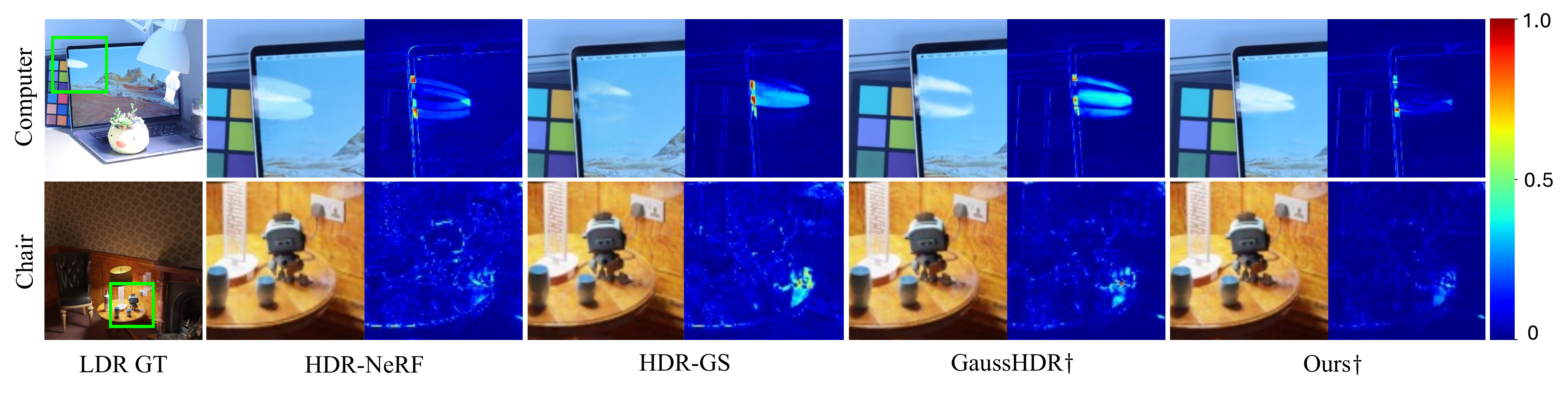}\vspace{-4pt}
    \caption{Qualitative comparisons on LDR views. For each method, we show the reconstructed LDR image and the residual map w.r.t. the ground truth. Competing methods exhibit noticeable missing content in saturated regions (e.g., screen reflections in the 1st row), indicating information loss after tone mapping, whereas our method effectively preserves fine structures and details.}\vspace{-2pt}
    \label{fig:ldr}
\end{figure*}

\begin{figure*}
    \centering
        \includegraphics[width=1\linewidth,clip, trim=20pt 30pt 20pt 18pt]{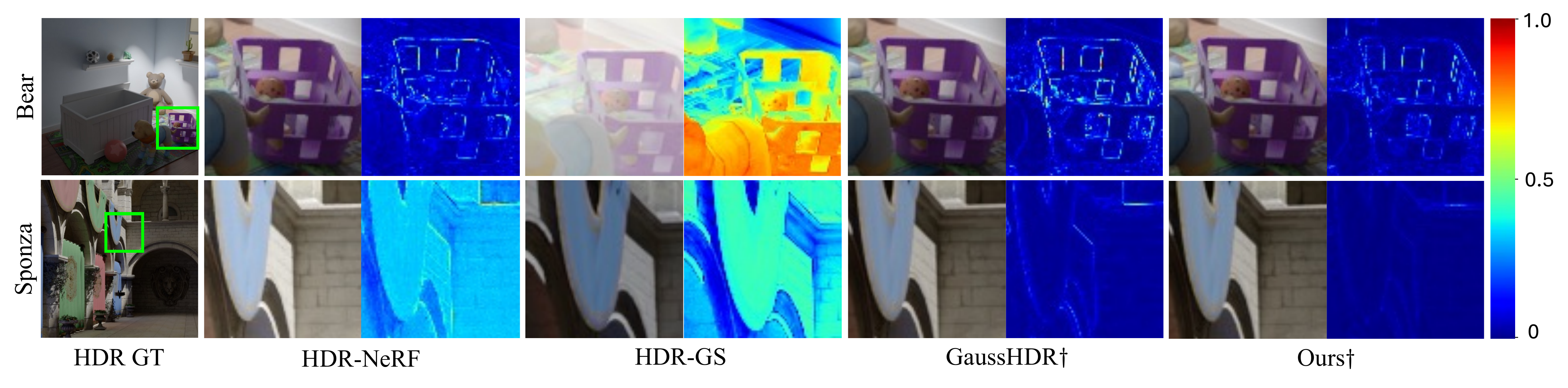}\vspace{-4pt}
    \caption{Qualitative comparisons on HDR views, where we include residual maps between the results and GT to highlight the difference. HDR-NeRF and HDR-GS struggle to reproduce correct illumination levels due to the absence of HDR supervision during training, leading to inaccurate brightness and lost details. By imposing cross-branch HDR consistency, our method faithfully estimates scene lighting and reconstructs fine structures (\eg, basket edges in the 1st row).}\vspace{-2pt}
    \label{fig:hdr}
\end{figure*}

\subsection{Quantitative Results}
Quantitative results on the realistic HDR-NeRF-Real and HDR-Plenoxels-Real datasets are included in \cref{tab:performance_real}. As can be seen, the proposed Scaffold-GS variant (\ie, Ours$\dagger$) achieves the overall best performance under both exposure settings,  yielding a PSNR gain of 0.59 dB over GaussHDR$\dagger$ on LDR-OE/exp3 of HDR-NeRF-Real. The 3DGS variant (\ie, Ours) also demonstrates competitive performance, showing even better performance than Scaffold-GS-based GaussHDR$\dagger$ on LDR-OE/exp1 of HDR-NeRF-Real.
Results on synthetic HDR-NeRF-Syn are reported in \cref{tab:performance_syn}, where Ours$\dagger$ shows consistent superiority over compared methods (\eg, a PSNR gain of 0.45 dB on  LDR-NE/exp3  over GaussHDR$\dagger$), achieving the overall leading performance.  It's worth noting that Ours$\dagger$ shows consistent superiority over all compared methods on the perceptual LPIPS across all benchmarks, while Ours also demonstrates competitive perceptual performance and surpasses GaussHDR on most metrics. This highlights the benefit of jointly modeling image exposure and ambient illumination for HDR-NVS, enabling both faithful LDR reconstruction and perceptually pleasing dynamic-range detail, particularly in challenging highlight regions (see \cref{sec:quali}). Detailed quantitative results of each scene are reported in the supplementary materials.

\subsection{Qualitative Results}\label{sec:quali}
Qualitative comparisons on LDR views are included in \cref{fig:ldr} and \cref{fig:hdr}, respectively. For each method, we also visualize the residual map with respect to the GT to intuitively demonstrate the difference. As can be seen, for LDR scenes, the compared methods exhibit noticeable missing contents for saturated regions (\eg, the reflection region in \cref{fig:ldr} ), indicating unconstrained information loss after tone mapping. In contrast, our method effectively preserves fine structures. We attribute this to the cross-branch HDR consistency loss that provides explicit supervision for regions of high signal intensity. For HDR scenes, methods such as HDR-NeRF and HDR-GS struggle to reproduce the correct illumination level, which results from the absence of HDR GT during training. While our method accurately estimates the lighting conditions by explicitly modeling exposure and ambient illumination, yielding effective reconstruction of details (\eg, edge of basket in the 1st row) over GaussHDR$\dagger$.  More qualitative comparisons are included in the supplementary materials.

\begin{table}[t]
\centering
% \Large
\caption{Efficiency analysis measured on a single NVIDIA A6000 GPU. Rendering time and throughput are evaluated at an output resolution of $400 \times 400$. } \vspace{-2pt}
\resizebox{1\linewidth}{!}{%
\begin{tabular}{l|cc|cc}
\toprule
 Method &  \makecell{Rendering \\(ms)} &  \makecell{Throughput\\(FPS)}  & \makecell{Training \\(min)} &  \makecell{Memory \\(MB)}  \\
\midrule
HDR-NeRF                &  4189	&  0.24 &   500  &  11049  \\
HDR-GS             &  9	  &  117 &     10  &  5014 \\
GaussHDR             & 19	&53         &  28 & 5596 \\
GaussHDR$\dagger$          &  26	&38        &  21   &   6221 \\
Ours       &  13	& 76    & 15    & 3274 \\
Ours$\dagger$   & 19	& 53     & 18  &   3920 \\
\bottomrule
\end{tabular}
} \vspace{-4pt}
\label{tab:efficiency}
\end{table}

\begin{table*}[t]
\centering
  \caption{Ablation studies on HDR-NeRF-Real and HDR-Plenoxels-Real dataset, with the best and second-best results highlighted in \colorbox{red!30}{red} and \colorbox{yellow!30}{yellow}.  IE branch indicates the baseline containing only the image-exposure branch. GI branch, HDR-cons and I-GS denote Gaussian-illumination branch, self-consistent HDR learning and illumination-guided gradient scaling, respectively.}
\label{tab:ablation}
\Large
\resizebox{1\linewidth}{!}{
\begin{tabular}{l|ccc ccc|ccc ccc}
\toprule
\multirow{3}{*}{Method}& \multicolumn{6}{c|}{HDR-NeRF-Real} &\multicolumn{6}{c}{HDR-Plenoxels-Real}\\
\cmidrule(lr){2-7} \cmidrule(lr){8-13} 
   & \multicolumn{3}{c}{LDR-OE ( $t_1, t_3, t_5$ )} & \multicolumn{3}{c|}{LDR-NE ( $t_2, t_4$ )} & \multicolumn{3}{c}{LDR-OE ( $t_1, t_3, t_5$ )} & \multicolumn{3}{c}{LDR-NE ( $t_2, t_4$ )} \\
\cmidrule(lr){2-4} \cmidrule(lr){5-7} \cmidrule(lr){8-10} \cmidrule(lr){11-13}
 & PSNR $\uparrow$ & SSIM $\uparrow$ & LPIPS $\downarrow$ & PSNR $\uparrow$ & SSIM $\uparrow$ & LPIPS $\downarrow$ & PSNR $\uparrow$ & SSIM $\uparrow$ & LPIPS $\downarrow$ & PSNR $\uparrow$ & SSIM $\uparrow$ & LPIPS $\downarrow$ \\
\midrule 
IE branch & 36.18 & 0.9766 & 0.0105 & 33.38 & 0.9727 & 0.0142 & 32.73 & 0.9576 & 0.0274 & \cellcolor{yellow!30}29.47 & \cellcolor{yellow!30}0.9368 & 0.0358\\
+ GI branch&36.27 & 0.9770 & 0.0096 & 33.46 & 0.9728 & 0.0132 &32.83 & 0.9570 & 0.0271 & 29.34 & 0.9350 & 0.0362 \\
+ HDR-cons &\cellcolor{yellow!30}36.43 & \cellcolor{yellow!30}0.9774 & \cellcolor{yellow!30}0.0094 & \cellcolor{yellow!30}33.84 & \cellcolor{yellow!30}0.9732 & \cellcolor{yellow!30}0.0129 & \cellcolor{yellow!30}32.97 & \cellcolor{yellow!30}0.9584 & \cellcolor{yellow!30}0.0262 & 29.45 & 0.9364 & \cellcolor{yellow!30}0.0354\\
+ I-GS &\cellcolor{red!30}36.91 & \cellcolor{red!30}0.9777 & \cellcolor{red!30}0.0093 & \cellcolor{red!30}34.15 & \cellcolor{red!30}0.9737 & \cellcolor{red!30}0.0124 &\cellcolor{red!30}33.06 & \cellcolor{red!30}0.9592 & \cellcolor{red!30}0.0253 & 2\cellcolor{red!30}9.76 & \cellcolor{red!30}0.9389 & \cellcolor{red!30}0.0339\\ 
\bottomrule
\end{tabular}}\vspace{-4pt}
\end{table*}

\subsection{Efficiency Analysis}
We conduct a comprehensive efficiency comparison in \cref{tab:efficiency},  reporting rendering speed, throughput, training time and GPU memory usage.  Despite the additional radiance composition step, our method remains efficient.  Ours and Ours$\dagger$ achieve real-time throughput of 76 FPS and 53 FPS at a target resolution of $400\times 400$, which is $322\times$ and $220\times$ faster than HDR-NeRF, respectively.  Compared with efficient HDR-GS, our method provides promising performance improvement (\eg,  3.98 dB PSNR gain on HDR-NeRF-Syn/exp3) with tolerable reduction in speed.  Compared with well-performing GaussHDR, our models deliver faster rendering speed (\eg, Ours is 1.43$\times$ faster than GaussHDR) and lower GPU memory consumption, while still offering overall better reconstruction performance.

\begin{figure}[t]
    \centering
    \includegraphics[width=1.02\linewidth,clip, trim=10pt 14pt 8pt 8pt]{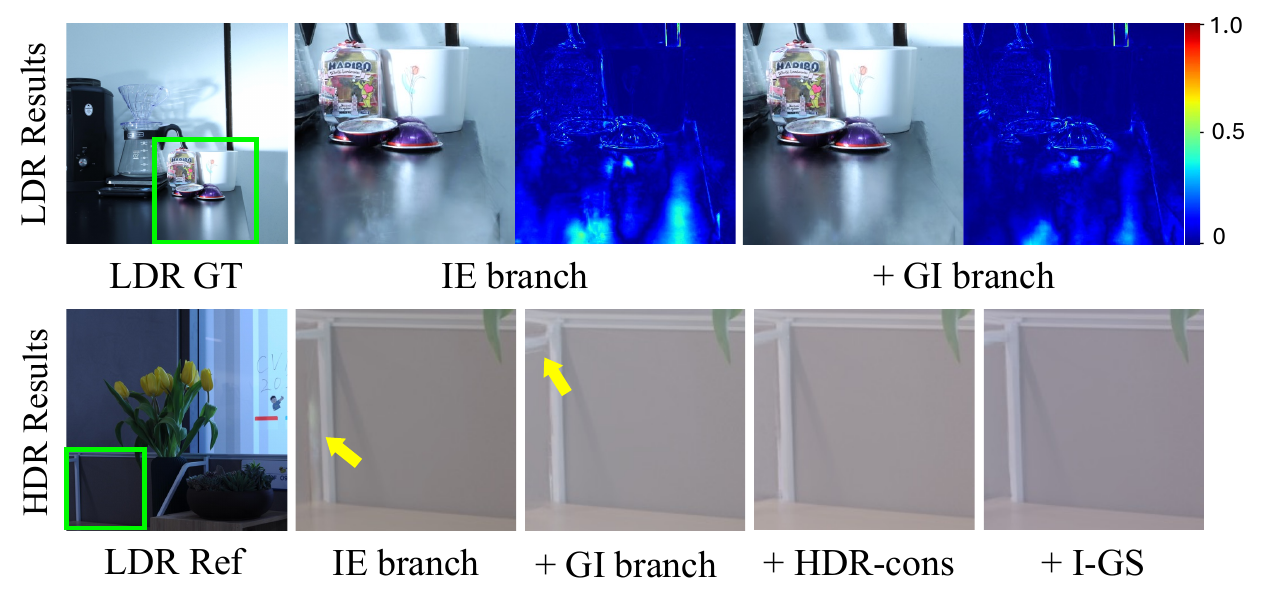}\vspace{-2pt}
    \caption{Visualization of ablation studies, where residual maps between LDR results and LDR GT are provided. Including GI branch effectively captures lighting-dependent appearance (\eg, table reflections in 1st row) and reduces color distortions. Introducing HDR-cons and I-GS further refines structural details.}\vspace{-2pt}
    \label{fig:ablation}
\end{figure}

\subsection{Ablation Studies}
We conduct the ablation studies starting from a baseline that contains the IE branch only (denoted as \textit{IE branch}).  Then we progressively equip this baseline with the GI branch, cross-branch HDR consistency loss, and illumination-guided gradient scaling to evaluate the contribution of each component. All ablations are conducted under the exposure setting of exp3 across HDR-NeRF-Real and HDR-Plenoxels-Real datasets for generality.

\noindent \textbf{Gaussian-illumination branch.} As shown in \cref{tab:ablation}, introducing Gaussian-illumination branch (denoted as \textit{+ GI branch}) yields consistent gains on HDR-NeRF-Real and overall improvement on HDR-Plenoxels-Real. We attribute this to the enhanced ability of the full framework to capture both exposure-scaled and lighting-conditioned appearance variations. As shown \cref{fig:ablation}, including the GI branch effectively improves the modeling of the lighting-dependent appearance (\eg, table surface reflection in the 1st row) and corrects texture distortion (\eg, desk shelf in the 2nd row).

\noindent \textbf{Self-consistent HDR learning.}
We verify the effectiveness of self-consistent HDR learning by imposing the cross-branch HDR consistency loss $\mathcal{L}_{cons}$ between IE and GI branches (denoted as \textit{+ HDR-cons}). As shown in \cref{tab:ablation}, imposing consistency loss leads to noticeable improvement, with a maximum PSNR gain of 0.38 dB on HDR-NeRF-Real.  Qualitative results in \cref{fig:ablation} indicate that self-consistent HDR learning helps to refine texture distortions.

\noindent \textbf{Illumination-guided gradient scaling.} We further include  illumination-guided gradient scaling  (denoted as \textit{+ I-GS}) during training to verify its effectiveness. As can be seen 
  As can be seen from \cref{fig:ablation}, scaling the per-Gaussian gradient with the proposed I-GS strategy introduces significant improvement, with a maximum PSNR gain of 0.48 dB and 0.31 dB on HDR-NeRF-Real and HDR-Plenoxels-Real, yielding the best overall configuration. Visualization in \cref{fig:ablation} shows that I-GS effectively alleviates texture distortions in shadow regions by protecting Gaussians from insufficient splitting.

\section{Conclusion}\label{sec:conclu} 
 In this paper, we propose a physically inspired HDR-NVS framework.   To reflect the illumination-dependent appearance, we decompose the Gaussian color into intrinsic reflectance and adjustable ambient illumination. An image-exposure branch and a Gaussian-illumination branch jointly model complementary dynamic range details. An HDR consistency loss is imposed across these branches to enable self-supervision for HDR contents.  The proposed gradient scaling strategy further amplifies the Gaussian gradients to prevent under-densified representations. Extensive experiments show that our method effectively reconstructs dynamic range details, while keeping the real-time rendering speed. Ablation studies further verify the effectiveness of each proposed component in our framework.
\appendix
\clearpage
\setcounter{page}{1}
\maketitlesupplementary

\renewcommand{\algorithmicrequire}{ \textbf{Input:}}     %Use Input in the format of Algorithm
\renewcommand{\algorithmicensure}{ \textbf{Output:}}    %UseOutput in the format of Algorithm
\definecolor{lightgrey}{RGB}{211,211,211}

\noindent This supplementary document is organized as follows:

– Section~\ref{sec:procedure} provides a detailed explanation and pseudo-code to clarify the training procedure of the proposed PhysHDR-GS.

– Section~\ref{sec:quanti} reports per-scene quantitative results on all adopted datasets.

– Section~\ref{sec:qualitative} includes more qualitative comparisons on both LDR and HDR views.

– Section~\ref{sec:setting} introduces further experimental details, including dataset preparation and implementation settings.

 – Section~\ref{sec:discussion} includes our observation for HDR-NVS under low-light conditions and discusses potential solutions to address such scenarios.

\begin{algorithm*}[!t] 
\caption{Training Procedure of PhysHDR-GS}\label{code:training}
\textbf{Input}: Reflectance $H_r$, illumination $L_a$, opacity $\{\boldsymbol{\alpha}_i\}_{i=1}^N$, centers $\{\boldsymbol{\mu}_i\}_{i=1}^N$, covariances $\{\boldsymbol{\Sigma}_i\}_{i=1}^N$, input LDR views $\{I_{gt}^k\}_{k=1}^K$, lighting level $\{l^k\}_{k=1}^K$, exposure $\{t^k\}_{k=1}^K$ \\
\textbf{Output}: Optimized HDR 3D Gaussian primitives $G^{3D}_{HDR}$ \\
\begin{algorithmic}[1]
\STATE Initialize $H_r$, $L_a$
\WHILE{iteration $\leq$ MaxIteration}
\STATE $I_{gt}, l, t \gets \text{SampleTrainingView}(\{I_{gt}^k\}_{k=1}^K, \{l^k\}_{k=1}^K, \{t^k\}_{k=1}^K)$;

\STATE \hfill \textit{/* Sec.~4.1: Physical Radiance Composition */}

\STATE $\boldsymbol{c} = g(L_a, H_r)$;  \hfill \textit{/* Image-exposure (IE) branch */}

\STATE $G^{3D}_{HDR} = \{(\boldsymbol{\mu}_i,\boldsymbol{\Sigma}_i,\boldsymbol{\alpha}_i,\boldsymbol{c}_i)\}_{i=1}^{N}$; 

\STATE $I_{HDR}$ $\gets$ Rasterize($G^{3D}_{HDR}$);

\STATE $I_{HDR} \times t \gets$ RescaleExposure($I_{HDR}$, $t$);  

\STATE \hfill \textit{/* Gaussian-illumination (GI) branch */}
\STATE $\hat{L}_a = \varphi(L_a, l)$; \hfill \textit{/*  Virtual illumination */}   
\STATE $\hat{\boldsymbol{c}} = g(\hat{L}_a, H_r)$;
\STATE $\hat{G}^{3D}_{HDR} = \{(\boldsymbol{\mu}_i,\boldsymbol{\Sigma}_i,\boldsymbol{\alpha}_i,\hat{\boldsymbol{c}}_i)\}_{i=1}^{N}$; \hfill \textit{/* Relit Gaussians */}  
\STATE $\hat{I}_{HDR}$ $\gets$ Rasterize($\hat{G}^{3D}_{HDR}$);

\STATE \hfill \textit{/* Sec.~4.2: Self-Consistent HDR Fusion */}

\STATE $\{I^{IG}_{LDR}, I^{GI}_{LDR},I_{LDR}\}
\gets \text{ToneMapping}(f, I_{HDR}\times t, \hat{I}_{HDR})$;\hfill \textit{/* Tone-mapped LDR learning */}
\STATE $\mathcal{L}_{rec} \gets \text{ReconstructionLoss}(\{I_{LDR}, I^{IG}_{LDR}, I^{GI}_{LDR}\}, I_{gt})$;

\STATE $\mathcal{L}_{cons} \gets \text{HDRConsistencyLoss}(I_{HDR}^{IE}, \hat{I}_{HDR})$;\hfill \textit{/* Self-consistent HDR learning */}

\STATE $\mathcal{L}_{total} \gets \text{WeightedLossSum}(\mathcal{L}_{rec}, \mathcal{L}_{cons}, \mathcal{L}_{unit})$;
\STATE  $L_a$, $H_r$,  $\boldsymbol{\alpha}$, $\boldsymbol{c}$, $\boldsymbol{\mu}$, $\boldsymbol{\Sigma}$ $\gets$ AdamOptimize($\nabla\mathcal{L}_{total}$)

\STATE \hfill \textit{/* Sec.~4.3: Illumination-Guided Gradient Scaling */}

\STATE $s_a \gets \text{GradientScaling} (s, L_a, \hat{L}_a$); 
\IF {$ s_a \nabla\mathcal{L}_{total} > \tau_p$}
\STATE Densification;
\ENDIF
\ENDWHILE
\end{algorithmic}
\end{algorithm*}

\vspace{-2pt}\section{Training Procedure}\label{sec:procedure}\vspace{-2pt}
The pseudo-code of the training procedure is summarized in Algorithm~\ref{code:training}. PhysHDR-GS is built from three components: physical radiance composition, self-consistent HDR fusion and illumination-guided gradient scaling. Physical radiance modeling incorporates an image–exposure (IE) branch and a Gaussian–illumination (GI) branch. The IE branch composes intrinsic reflectance and ambient illumination into per-Gaussian color and synthesizes the HDR Gaussian primitives, while the GI branch adjusts the illumination and produces relit HDR Gaussian primitives. At the self-consistent HDR fusion stage, the projected HDR image from the IE branch is re-exposed to match the brightness of the relit HDR image of the GI branch. These HDR intermediates are tone-mapped and fused as the final LDR output. During training,  the cross-branch HDR consistency loss and illumination-guided gradient scaling are adopted to enable self-supervision for HDR learning and alleviate under-densified representations in over-/under-exposed regions, respectively.

\begin{table*}[t]
\centering
\caption{Per-scene quantitative comparisons of each scene on HDR-NeRF-Real~\cite{huang2022hdr} dataset. For each scene, the best and second-best results are highlighted in \colorbox{red!30}{red} and \colorbox{yellow!30}{yellow}. LDR-OE and LDR-NE denote the LDR results with exposure $\{t_1,t_3,t_5\}$ and $\{t_2,t_4\}$, respectively.  }
\label{tab:hdrnerfreal_per_scene}
\Large
\resizebox{\linewidth}{!}{
\begin{tabular}{c|c|l|ccc ccc ccc ccc}
\toprule
 \multicolumn{3}{c|}{\multirow{2}{*}{Method}}
& \multicolumn{3}{c}{\textit{Box}}
& \multicolumn{3}{c}{\textit{Computer}}
& \multicolumn{3}{c}{\textit{Flower}}
& \multicolumn{3}{c}{\textit{Luckycat}} \\
\cmidrule(lr){4-6}\cmidrule(lr){7-9}\cmidrule(lr){10-12}\cmidrule(lr){13-15}
\multicolumn{3}{c|}{}& PSNR$\uparrow$ & SSIM$\uparrow$ & LPIPS$\downarrow$
& PSNR$\uparrow$ & SSIM$\uparrow$ & LPIPS$\downarrow$
& PSNR$\uparrow$ & SSIM$\uparrow$ & LPIPS$\downarrow$
& PSNR$\uparrow$ & SSIM$\uparrow$ & LPIPS$\downarrow$ \\
\midrule
\multirow{12}{*}{\rotatebox{90}{exp3}}
& \multirow{6}{*}{\rotatebox{90}{LDR-OE}} 
& HDR-NeRF~\cite{huang2022hdr}
& 35.13 & 0.9612 & 0.055
&  34.28& 0.9486 & 0.076
&   33.21 & 0.9525 & 0.058 
&   34.47 & 0.9504 & 0.063  \\
& & HDR-GS~\cite{cai2024hdr}
&  36.21 & 0.9797 & 0.011
& 36.02 & 0.9720 & 0.017
&  32.44 & 0.9603 & 0.039
&  34.82 & 0.9668 & 0.018   \\
& & GaussHDR~\cite{liu2025gausshdr}
&  37.17 & 0.9821 & 0.009
&  36.24 & 0.9735 & 0.015
&  34.61 & 0.9681 & 0.020
&  36.19 & 0.9718 & 0.014 \\
& & GaussHDR$\dagger$~\cite{liu2025gausshdr}
&  37.18 & 0.9834 & 0.008 
&  \cellcolor{yellow!30}36.37 & 0.9737 & 0.013
&  \cellcolor{yellow!30}35.06 & \cellcolor{yellow!30}0.9757 & \cellcolor{yellow!30}0.012
&  36.67 & 0.9750 & 0.012\\
& & Ours
&   \cellcolor{red!30}37.94 & \cellcolor{red!30}0.9845 & \cellcolor{yellow!30}0.007
& 35.67 & \cellcolor{red!30}0.9761 & \cellcolor{yellow!30}0.012
&33.91 & 0.9732 & 0.015
&37.23 & \cellcolor{red!30}0.9779 & \cellcolor{yellow!30}0.010\\
& & Ours$\dagger$
&  \cellcolor{yellow!30}37.88 & \cellcolor{yellow!30}0.9836 & \cellcolor{red!30}0.007
&\cellcolor{red!30}37.12 & \cellcolor{yellow!30}0.9746 & \cellcolor{red!30}0.011
&\cellcolor{red!30}35.32 & \cellcolor{red!30}0.9764 & \cellcolor{red!30}0.011
&\cellcolor{red!30}37.32 & \cellcolor{yellow!30}0.9760 & \cellcolor{red!30}0.009\\
\cmidrule{2-15}
& \multirow{6}{*}{\rotatebox{90}{LDR-NE}} 
& HDR-NeRF~\cite{huang2022hdr}
&  31.20 & 0.9518 & 0.081
&33.56 & 0.9487 & 0.080
&30.05 & 0.9439 & 0.072
&33.81 & 0.9458 & 0.063 \\
& & HDR-GS~\cite{cai2024hdr}
&  29.04 & 0.9697 & 0.024
& 33.23 & 0.9696 & 0.021
&29.08 & 0.9539 & 0.049
&32.75 & 0.9612 & 0.021  \\
& & GaussHDR
&33.18 & 0.9774 & 0.013
&35.12 & 0.9716 & 0.018
&30.94 & 0.9648 & 0.024
&34.72 & 0.9690 & 0.015 \\
& & GaussHDR$\dagger$~\cite{liu2025gausshdr}
& 33.19 & \cellcolor{yellow!30}0.9784 & 0.011
&\cellcolor{yellow!30}35.37 & 0.9715 & 0.015
&\cellcolor{red!30}31.71 & \cellcolor{yellow!30}0.9730 & \cellcolor{yellow!30}0.014
&35.10 & \cellcolor{yellow!30}0.9722 & 0.013 \\
& & Ours
& \cellcolor{red!30}33.75 & \cellcolor{red!30}0.9794 & \cellcolor{yellow!30}0.011
&34.53 & \cellcolor{red!30}0.9738 & \cellcolor{yellow!30}0.014
&30.51 & 0.9671 & 0.020
&\cellcolor{red!30}35.92 & \cellcolor{red!30}0.9747 & \cellcolor{yellow!30}0.011 \\
& & Ours$\dagger$
&  \cellcolor{yellow!30}33.42 &  0.9779 & \cellcolor{red!30}0.011
&\cellcolor{red!30}35.87 & \cellcolor{yellow!30}0.9721 & \cellcolor{red!30}0.013
&\cellcolor{yellow!30}31.68 & \cellcolor{red!30}0.9730 & \cellcolor{red!30}0.014
&\cellcolor{yellow!30}35.62 & 0.9719 & \cellcolor{red!30}0.011 \\
\midrule
\multirow{12}{*}{\rotatebox{90}{exp1}}
& \multirow{6}{*}{\rotatebox{90}{LDR-OE}} 
& HDR-NeRF~\cite{huang2022hdr}
&  35.13 & 0.9612 & 0.055
&34.26 & 0.9486 & 0.076
&33.18 & 0.9525 & 0.058
&34.46 & 0.9504 & 0.063 \\
& & HDR-GS~\cite{cai2024hdr}
&  33.63 & 0.9692 & 0.019
&34.50 & 0.9672 & 0.020
&30.14 & 0.9425 & 0.051
&33.56 & 0.9597 & 0.022\\
& & GaussHDR~\cite{liu2025gausshdr}
&  \cellcolor{yellow!30}36.21 & \cellcolor{yellow!30}0.9780 & 0.012
&\cellcolor{yellow!30}34.85 & 0.9679 & 0.019
&32.78 & 0.9617 & 0.024
&34.52 & 0.9637 & 0.018 \\
& & GaussHDR$\dagger$~\cite{liu2025gausshdr}
&  35.44 & 0.9780 & 0.010
&34.38 & 0.9679 & 0.016
&\cellcolor{yellow!30}33.23 & \cellcolor{red!30}0.9695 & \cellcolor{yellow!30}0.015
&\cellcolor{red!30}35.46 & \cellcolor{yellow!30}0.9687 & 0.015\\
& & Ours
&  \cellcolor{red!30}36.55 & \cellcolor{red!30}0.9800 & \cellcolor{red!30}0.009
&34.48 & \cellcolor{yellow!30}0.9692 & \cellcolor{yellow!30}0.016
&32.43 & 0.9638 & 0.022
&35.10 & \cellcolor{red!30}0.9699 & \cellcolor{yellow!30}0.014 \\
& & Ours$\dagger$
&  35.69 & 0.9766 & \cellcolor{yellow!30}0.010
&\cellcolor{red!30}35.07 & \cellcolor{red!30}0.9693 & \cellcolor{red!30}0.013
&\cellcolor{red!30}33.42 & \cellcolor{yellow!30}0.9685 & \cellcolor{red!30}0.014
&\cellcolor{yellow!30}35.17 & 0.9677 & \cellcolor{red!30}0.013  \\
\cmidrule{2-15}     
& \multirow{6}{*}{\rotatebox{90}{LDR-NE}} 
& HDR-NeRF~\cite{huang2022hdr}
&   31.17 & 0.9518 & 0.081
&31.17 & 0.9518 & 0.081
&30.05 & 0.9439 & 0.072
&33.80 & 0.9457 & 0.063 \\
& & HDR-GS~\cite{cai2024hdr}
&   27.35 & 0.9578 & 0.033
&31.20 & 0.9643 & 0.025
&28.61 & 0.9403 & 0.055
&31.47 & 0.9516 & 0.027  \\
& & GaussHDR~\cite{liu2025gausshdr}
&  \cellcolor{red!30}33.18 & \cellcolor{yellow!30}0.9742 & 0.014
&\cellcolor{yellow!30}34.31 & 0.9671 & 0.021
&30.86 & 0.9605 & 0.026
&33.56 & 0.9593 & 0.019\\
& & GaussHDR$\dagger$~\cite{liu2025gausshdr}
&   32.90 & 0.9741 & 0.013
&34.04 & 0.9670 & 0.018
&\cellcolor{yellow!30}31.90 & \cellcolor{red!30}0.9677 & \cellcolor{yellow!30}0.016
&\cellcolor{red!30}34.32 & \cellcolor{yellow!30}0.9646 & 0.017 \\
& & Ours
&  32.33 & \cellcolor{red!30}0.9746 & \cellcolor{yellow!30}0.013
&33.60 & \cellcolor{yellow!30}0.9677 & \cellcolor{yellow!30}0.018
&30.55 & 0.9605 & 0.026
&\cellcolor{yellow!30}34.06 & \cellcolor{red!30}0.9660 & \cellcolor{yellow!30}0.015 \\
& & Ours$\dagger$
&  \cellcolor{yellow!30}33.02 & 0.9730 & \cellcolor{red!30}0.013
&\cellcolor{red!30}34.55 & \cellcolor{red!30}0.9683 & \cellcolor{red!30}0.015
&\cellcolor{red!30}32.13 & \cellcolor{yellow!30}0.9665 & \cellcolor{red!30}0.015
&33.92 & 0.9630 & \cellcolor{red!30}0.015 \\
\bottomrule
\end{tabular}}
\end{table*}

\begin{table*}[t]
  \centering
  \caption{Per-scene quantitative comparisons on the HDR-Plenoxels-Real~\cite{jun2022hdr} dataset. HDR-NeRF~\cite{huang2022hdr} is not included, as it is evaluated on HDR-Plenoxels-Real. For each scene, the best and second-best results are highlighted in \colorbox{red!30}{red} and \colorbox{yellow!30}{yellow}. LDR-OE and LDR-NE denote LDR results with exposure sets $\{t_1,t_3,t_5\}$ and $\{t_2,t_4\}$, respectively.}
  \label{tab:hdrplenreal_per_scene}
  \Large
  \resizebox{1\linewidth}{!}{
  \begin{tabular}{c|c|l|ccc ccc ccc ccc}
  \toprule
   \multicolumn{3}{c|}{\multirow{2}{*}{Method}}
  & \multicolumn{3}{c}{\textit{Character}}
  & \multicolumn{3}{c}{\textit{Coffee}}
  & \multicolumn{3}{c}{\textit{Desk}}
  & \multicolumn{3}{c}{\textit{Plant}} \\
  \cmidrule(lr){4-6}\cmidrule(lr){7-9}\cmidrule(lr){10-12}\cmidrule(lr){13-15}
  \multicolumn{3}{c|}{}& PSNR$\uparrow$ & SSIM$\uparrow$ & LPIPS$\downarrow$
  & PSNR$\uparrow$ & SSIM$\uparrow$ & LPIPS$\downarrow$
  & PSNR$\uparrow$ & SSIM$\uparrow$ & LPIPS$\downarrow$
  & PSNR$\uparrow$ & SSIM$\uparrow$ & LPIPS$\downarrow$ \\
  \midrule
  \multirow{10}{*}{\rotatebox{90}{exp3}} & \multirow{5}{*}{\rotatebox{90}{LDR-OE}} 
  & HDR-GS~\cite{cai2024hdr}
  &   36.08 & 0.9799 & 0.024
  & 28.13 & 0.9468 & 0.046
  & 30.07 & 0.9366 & 0.039
  & 30.39 & 0.9403 & 0.049\\
  & & GaussHDR~\cite{liu2025gausshdr}
  &   36.47 & 0.9813 & 0.024
  & 29.10 & 0.9512 & 0.041
  & 29.79 & 0.9355 & 0.038
  & 30.63 & 0.9439 & 0.043 \\
  & & GaussHDR$\dagger$~\cite{liu2025gausshdr}
  &  39.09 & 0.9847 & 0.017
  & \cellcolor{red!30}29.82 & \cellcolor{red!30}0.9597 & \cellcolor{yellow!30}0.029
  & \cellcolor{yellow!30}30.52 & \cellcolor{yellow!30}0.9417 & \cellcolor{yellow!30}0.031
  & \cellcolor{red!30}32.05 & \cellcolor{red!30}0.9499 & \cellcolor{yellow!30}0.036 \\
  & & Ours
  &  \cellcolor{yellow!30}39.45 & \cellcolor{yellow!30}0.9854 & \cellcolor{yellow!30}0.015
  & 28.94 & 0.9538 & 0.035
  & 29.79 & 0.9351 & 0.037
  & 30.56 & 0.9402 & 0.046 \\
  & & Ours$\dagger$
  &  \cellcolor{red!30}40.34 & \cellcolor{red!30}0.9870 & \cellcolor{red!30}0.013
  & \cellcolor{yellow!30}29.51 & \cellcolor{yellow!30}0.9585 & \cellcolor{red!30}0.026
  & \cellcolor{red!30}30.66 & \cellcolor{red!30}0.9428 & \cellcolor{red!30}0.028
  & \cellcolor{yellow!30}31.72 & \cellcolor{yellow!30}0.9486 & \cellcolor{red!30}0.034 \\
  \cmidrule{2-15}
  & \multirow{5}{*}{\rotatebox{90}{LDR-NE}} 
  & HDR-GS~\cite{cai2024hdr}
  &   --  &  --  &  -- 
  &  --  &  --  &  -- 
  & 28.12 & 0.9260 & 0.044
  & 29.09 & 0.9309 & 0.057  \\
  & & GaussHDR~\cite{liu2025gausshdr}
  &   --  &  --  &  -- 
  &  --  &  --  &  -- 
  & 28.26 & 0.9272 & 0.042
  & 29.59 & 0.9380 & 0.047 \\
  & &   GaussHDR$\dagger$~\cite{liu2025gausshdr}
  &   --  &  --  &  -- 
  &  --  &  --  &  -- 
  & \cellcolor{yellow!30}28.61 & \cellcolor{yellow!30}0.9330 & \cellcolor{yellow!30}0.034
  & \cellcolor{red!30}30.95 & \cellcolor{red!30}0.9451 & \cellcolor{yellow!30}0.040\\
  & & Ours
  &  --  &  --  &  -- 
  &  --  &  --  &  -- 
  & 28.12 & 0.9259 & 0.041
  & 29.47 & 0.9339 & 0.051\\
  & &  Ours$\dagger$
  &   --  &  --  &  --   
  &  --  &  --  &  --  
  & \cellcolor{red!30}28.91 & \cellcolor{red!30}0.9339 & \cellcolor{red!30}0.030
  & \cellcolor{yellow!30}30.61 & \cellcolor{yellow!30}0.9439 & \cellcolor{red!30}0.037 \\
  \midrule  
  \multirow{10}{*}{\rotatebox{90}{exp1}}
  & \multirow{5}{*}{\rotatebox{90}{LDR-OE}} 
  & HDR-GS~\cite{cai2024hdr}
  &  34.85 & 0.9773 & 0.029
  & 25.21 & 0.9194 & 0.070
  & 29.24 & 0.9250 & 0.044
  & 29.16 & 0.9275 & 0.059 \\
  & & GaussHDR~\cite{liu2025gausshdr}
  &  36.36 & 0.9801 & 0.024
  & 28.02 & 0.9432 & 0.048
  & 29.08 & 0.9242 & 0.044
  & 30.03 & 0.9355 & 0.048  \\
  & & GaussHDR$\dagger$~\cite{liu2025gausshdr}
  &  \cellcolor{yellow!30}38.24 & 0.9830 & 0.018
  & \cellcolor{yellow!30}29.31 & \cellcolor{yellow!30}0.9549 & \cellcolor{yellow!30}0.032
  & \cellcolor{red!30}29.91 & \cellcolor{red!30}0.9331 & \cellcolor{yellow!30}0.035
  & \cellcolor{red!30}31.48 & \cellcolor{red!30}0.9448 & \cellcolor{yellow!30}0.040 \\
  & & Ours
  &  38.13 & \cellcolor{yellow!30}0.9836 & \cellcolor{yellow!30}0.017
  & 28.69 & 0.9479 & 0.040
  & 28.78 & 0.9240 & 0.040
  & 29.19 & 0.9301 & 0.050 \\
  & & Ours$\dagger$
  &  \cellcolor{red!30}38.68 & \cellcolor{red!30}0.9839 & \cellcolor{red!30}0.015
  & \cellcolor{red!30}29.54 & \cellcolor{red!30}0.9551 & \cellcolor{red!30}0.029
  & \cellcolor{yellow!30}29.87 & \cellcolor{yellow!30}0.9326 & \cellcolor{red!30}0.035
  & \cellcolor{yellow!30}31.26 & \cellcolor{yellow!30}0.9444 & \cellcolor{red!30}0.037\\
  \cmidrule{2-15}     
  & \multirow{5}{*}{\rotatebox{90}{LDR-NE}} 
  & HDR-GS~\cite{cai2024hdr}
  &    --  &  --  &  -- 
  &  --  &  --  &  -- 
  & 27.12 & 0.9134 & 0.052
  & 27.52 & 0.9167 & 0.069  \\
  & & GaussHDR~\cite{liu2025gausshdr}
  &   --  &  --  &  -- 
  &  --  &  --  &  -- 
  & \cellcolor{yellow!30}27.56 & 0.9163 & 0.046
  & 28.96 & 0.9297 & 0.053  \\
  & & GaussHDR$\dagger$~\cite{liu2025gausshdr}
  &   --  &  --  &  -- 
  &  --  &  --  &  -- 
  & 27.21 & \cellcolor{yellow!30}0.9225 & \cellcolor{yellow!30}0.038
  & \cellcolor{red!30}30.55 & \cellcolor{red!30}0.9405 & \cellcolor{yellow!30}0.044  \\
  & & Ours
  &    --  &  --  &  -- 
  &  --  &  --  &  -- 
  & 27.37 & 0.9152 & 0.042
  & 27.82 & 0.9234 & 0.053\\
  & & Ours$\dagger$
  &  --  &  --  &  -- 
  &  --  &  --  &  -- 
  & \cellcolor{red!30}27.96 & \cellcolor{red!30}0.9227 & \cellcolor{red!30}0.037
  & \cellcolor{yellow!30}30.17 & \cellcolor{yellow!30}0.9399 & \cellcolor{red!30}0.040 \\
  \bottomrule
  \end{tabular}}
  \end{table*}

\begin{table*}[t]
\centering
\caption{Per-scene quantitative comparisons on the HDR-NeRF-Syn~\cite{huang2022hdr} dataset under the exp3 exposure setting. For each scene, the best and second-best results are highlighted in \colorbox{red!30}{red} and \colorbox{yellow!30}{yellow}. LDR-OE and LDR-NE denote LDR results with exposure sets $\{t_1,t_3,t_5\}$ and $\{t_2,t_4\}$, respectively. }
\label{tab:hdrnerfsyn_per_sceneexp3}
\Large
\resizebox{1\linewidth}{!}{
\begin{tabular}{c|l|ccc ccc ccc ccc}
\toprule
 \multicolumn{2}{c|}{\multirow{2}{*}{Method}}
& \multicolumn{3}{c}{\textit{Bathroom}}
& \multicolumn{3}{c}{\textit{Bear}}
& \multicolumn{3}{c}{\textit{Chair}}
& \multicolumn{3}{c}{\textit{Desk}} \\
\cmidrule(lr){3-5}\cmidrule(lr){6-8}\cmidrule(lr){9-11}\cmidrule(lr){12-14}
\multicolumn{2}{c|}{}& PSNR$\uparrow$ & SSIM$\uparrow$ & LPIPS$\downarrow$
& PSNR$\uparrow$ & SSIM$\uparrow$ & LPIPS$\downarrow$
& PSNR$\uparrow$ & SSIM$\uparrow$ & LPIPS$\downarrow$
& PSNR$\uparrow$ & SSIM$\uparrow$ & LPIPS$\downarrow$ \\
\midrule
 \multirow{6}{*}{\rotatebox{90}{LDR-OE}} 
& HDR-NeRF~\cite{huang2022hdr}
& 33.98 & 0.9067 & 0.108
& 43.96 & 0.9917 & 0.007
& 34.36 & 0.9286 & 0.064
& 38.78 & 0.9735 & 0.021  \\
& HDR-GS~\cite{cai2024hdr}
&   42.29 & 0.9796 & 0.008
& 38.61 & 0.9835 & 0.015
& 37.75 & 0.9734 & 0.013
& 43.62 & 0.9885 & 0.004  \\
 & GaussHDR~\cite{liu2025gausshdr}
& 42.38 & 0.9803 & 0.006
& 45.60 & 0.9934 & 0.002
& 37.44 & 0.9671 & 0.015
& 42.84 & 0.9888 & 0.004 \\
 & GaussHDR$\dagger$~\cite{liu2025gausshdr}
&   \cellcolor{yellow!30}42.97 & \cellcolor{red!30}0.9844 & 0.006
& \cellcolor{yellow!30}46.71 & \cellcolor{yellow!30}0.9945 & \cellcolor{yellow!30}0.001
& \cellcolor{yellow!30}38.48 & \cellcolor{red!30}0.9746 & \cellcolor{yellow!30}0.011
& \cellcolor{yellow!30}43.77 & \cellcolor{yellow!30}0.9905 & 0.003 \\
& Ours
& 42.34 & 0.9790 & \cellcolor{yellow!30}0.006
& 46.76 & 0.9942 & 0.002
& 37.89 & 0.9708 & 0.012
& 43.78 & 0.9881 & \cellcolor{yellow!30}0.003 \\
 & Ours$\dagger$
& \cellcolor{red!30}43.25 & \cellcolor{yellow!30}0.9839 & \cellcolor{red!30}0.005
& \cellcolor{red!30}47.14 & \cellcolor{red!30}0.9945 & \cellcolor{red!30}0.001
& \cellcolor{red!30}38.85 & \cellcolor{yellow!30}0.9745 & \cellcolor{red!30}0.010
& \cellcolor{red!30}44.50 & \cellcolor{red!30}0.9910 & \cellcolor{red!30}0.002\\
\midrule
 \multirow{6}{*}{\rotatebox{90}{LDR-NE}} 
& HDR-NeRF~\cite{huang2022hdr}
&  33.30 & 0.9041 & 0.115
& 42.46 & 0.9906 & 0.009
& 33.55 & 0.9214 & 0.072
& 39.47 & 0.9748 & 0.019 \\
& HDR-GS~\cite{cai2024hdr}
&   37.81 & 0.9745 & 0.024
& 12.70 & 0.7217 & 0.196
& 32.73 & 0.9561 & 0.040
& 37.24 & 0.9815 & 0.014  \\
 & GaussHDR~\cite{liu2025gausshdr}
& 42.04 & 0.9813 & 0.006
& 44.92 & 0.9928 & 0.002
& 36.81 & 0.9651 & 0.016
& 42.45 & 0.9887 & 0.004 \\
 & GaussHDR$\dagger$~\cite{liu2025gausshdr}
&  \cellcolor{red!30}42.49 & \cellcolor{red!30}0.9849 & 0.006
& \cellcolor{yellow!30}45.51 & \cellcolor{yellow!30}0.9937 & 0.002
& \cellcolor{yellow!30}37.73 & \cellcolor{yellow!30}0.9729 & 0.013
& 43.12 & \cellcolor{yellow!30}0.9902 & 0.003 \\
& Ours
&  41.96 & 0.9801 & \cellcolor{yellow!30}0.006
& \cellcolor{red!30}45.80 & 0.9935 & \cellcolor{yellow!30}0.002
& 37.21 & 0.9693 & \cellcolor{yellow!30}0.013
& \cellcolor{red!30}43.31 & 0.9883 & \cellcolor{yellow!30}0.003\\
 & Ours$\dagger$
& \cellcolor{yellow!30}42.39 & \cellcolor{yellow!30}0.9844 & \cellcolor{red!30}0.005
& 45.40 & \cellcolor{red!30}0.9937 & \cellcolor{red!30}0.002
& \cellcolor{red!30}38.14 & \cellcolor{red!30}0.9733 & \cellcolor{red!30}0.011
& \cellcolor{yellow!30}43.12 & \cellcolor{red!30}0.9906 & \cellcolor{red!30}0.002\\
\midrule
 \multirow{6}{*}{\rotatebox{90}{HDR}} 
& HDR-NeRF~\cite{huang2022hdr}
&   23.06 & 0.9556 & 0.092
& 26.35 & 0.9631 & 0.024
& 26.38 & 0.9555 & 0.057
& 44.11 & 0.9936 & 0.007\\
& HDR-GS~\cite{cai2024hdr}
&    20.77 & 0.8493 & 0.095
& 9.10 & 0.6961 & 0.241
& 14.30 & 0.2937 & 0.404
& 29.40 & 0.8144 & 0.102 \\
 & GaussHDR~\cite{liu2025gausshdr}
& 35.35 & 0.9491 & 0.021
& 42.09 & 0.9879 & 0.006
& 37.15 & 0.9604 & 0.020
& 44.50 & 0.9932 & 0.006 \\
 & GaussHDR$\dagger$~\cite{liu2025gausshdr}
&  \cellcolor{yellow!30}36.45 & \cellcolor{red!30}0.9565 & 0.016
& \cellcolor{yellow!30}42.83 & \cellcolor{yellow!30}0.9891 & 0.004
& \cellcolor{yellow!30}38.42 & \cellcolor{yellow!30}0.9685 & 0.016
& \cellcolor{yellow!30}44.86 & 0.9933 & 0.005 \\
&  Ours
& 36.05 & 0.9495 & \cellcolor{yellow!30}0.016
& 42.61 & 0.9887 & \cellcolor{yellow!30}0.004
& 38.00 & 0.9655 & \cellcolor{yellow!30}0.016
& 44.83 & \cellcolor{yellow!30}0.9935 & \cellcolor{yellow!30}0.004 \\
 & Ours$\dagger$
& \cellcolor{red!30}36.63 & \cellcolor{yellow!30}0.9558 & \cellcolor{red!30}0.015
& \cellcolor{red!30}42.86 & \cellcolor{red!30}0.9891 & \cellcolor{red!30}0.004
& \cellcolor{red!30}38.55 & \cellcolor{red!30}0.9686 & \cellcolor{red!30}0.014
& \cellcolor{red!30}45.08 & \cellcolor{red!30}0.9938 & \cellcolor{red!30}0.004 \\
\midrule  
 \multicolumn{2}{c|}{\multirow{2}{*}{ }}
& \multicolumn{3}{c}{\textit{Diningroom}}
& \multicolumn{3}{c}{\textit{Dog}}
& \multicolumn{3}{c}{\textit{Sofa}}
& \multicolumn{3}{c}{\textit{Sponza}} \\
\cmidrule(lr){3-5}\cmidrule(lr){6-8}\cmidrule(lr){9-11}\cmidrule(lr){12-14}
\multicolumn{2}{c|}{}& PSNR$\uparrow$ & SSIM$\uparrow$ & LPIPS$\downarrow$
& PSNR$\uparrow$ & SSIM$\uparrow$ & LPIPS$\downarrow$
& PSNR$\uparrow$ & SSIM$\uparrow$ & LPIPS$\downarrow$
& PSNR$\uparrow$ & SSIM$\uparrow$ & LPIPS$\downarrow$ \\
\midrule
 \multirow{6}{*}{\rotatebox{90}{LDR-OE}} 
& HDR-NeRF~\cite{huang2022hdr}
&  43.33 & 0.9894 & 0.008
& 39.50 & 0.9840 & 0.013
& 39.83 & 0.9818 & 0.013
& 36.80 & 0.9701 & 0.020\\
& HDR-GS~\cite{cai2024hdr}
&   32.96 & 0.9254 & 0.090
& 41.65 & 0.9920 & 0.005
& 43.45 & 0.9921 & 0.003
& 41.96 & 0.9901 & 0.004 \\
 & GaussHDR~\cite{liu2025gausshdr}
&  40.44 & 0.9813 & 0.015
& 43.58 & 0.9927 & 0.003
& 43.82 & 0.9917 & 0.003
& 42.09 & 0.9868 & 0.008\\
 & GaussHDR$\dagger$~\cite{liu2025gausshdr}
&  \cellcolor{yellow!30}45.65 & \cellcolor{yellow!30}0.9948 & \cellcolor{yellow!30}0.002
& \cellcolor{yellow!30}44.59 & \cellcolor{yellow!30}0.9948 & 0.002
& \cellcolor{yellow!30}44.96 & \cellcolor{yellow!30}0.9934 & 0.002
& \cellcolor{red!30}43.80 & \cellcolor{red!30}0.9922 & \cellcolor{red!30}0.003\\
& Ours
& 44.62 & 0.9930 & 0.003
& 44.34 & 0.9939 & \cellcolor{yellow!30}0.002
& 44.66 & 0.9931 & \cellcolor{yellow!30}0.002
& 40.50 & 0.9660 & 0.034 \\
 & Ours$\dagger$
& \cellcolor{red!30}45.94 & \cellcolor{red!30}0.9949 & \cellcolor{red!30}0.001
& \cellcolor{red!30}45.18 & \cellcolor{red!30}0.9953 & \cellcolor{red!30}0.002
& \cellcolor{red!30}45.38 & \cellcolor{red!30}0.9942 & \cellcolor{red!30}0.002
& \cellcolor{yellow!30}43.85 & \cellcolor{yellow!30}0.9910 & \cellcolor{yellow!30}0.004\\
\midrule
 \multirow{6}{*}{\rotatebox{90}{LDR-NE}} 
& HDR-NeRF~\cite{huang2022hdr}
&  42.38 & 0.9892 & 0.008
& 38.47 & 0.9823 & 0.014
& 38.70 & 0.9806 & 0.013
& 36.23 & 0.9702 & 0.020 \\
& HDR-GS~\cite{cai2024hdr}
&     27.38 & 0.8968 & 0.135
& 15.10 & 0.7027 & 0.336
& 17.30 & 0.7788 & 0.258
& 36.28 & 0.9832 & 0.010\\
 & GaussHDR~\cite{liu2025gausshdr}
&  39.37 & 0.9795 & 0.017
& 42.71 & 0.9921 & 0.004
& 43.11 & 0.9918 & 0.003
& 41.76 & 0.9884 & 0.006\\
 & GaussHDR$\dagger$~\cite{liu2025gausshdr}
&  \cellcolor{yellow!30}44.26 & \cellcolor{yellow!30}0.9942 &  \cellcolor{yellow!30}0.002
& 42.13 & \cellcolor{yellow!30}0.9937 & 0.003
&  43.34 & 0.9930 & 0.003
& \cellcolor{red!30}43.34 & \cellcolor{red!30}0.9925 &  \cellcolor{red!30}0.003 \\
& Ours
&43.90 & 0.9927 & 0.003
& \cellcolor{yellow!30}43.31 & 0.9935 & \cellcolor{yellow!30}0.002
& \cellcolor{yellow!30}44.00 &  \cellcolor{yellow!30}0.9931 &  \cellcolor{yellow!30}0.002
& 40.36 & 0.9713 & 0.036\\
 & Ours$\dagger$
& \cellcolor{red!30}44.72 & \cellcolor{red!30}0.9944 & \cellcolor{red!30}0.002
&  \cellcolor{red!30}43.78 & \cellcolor{red!30}0.9949 & \cellcolor{red!30}0.002
& \cellcolor{red!30}44.62 & \cellcolor{red!30}0.9942 & \cellcolor{red!30}0.002
& \cellcolor{yellow!30}43.37 &  \cellcolor{yellow!30}0.9914 &  \cellcolor{yellow!30}0.004\\
\midrule
 \multirow{6}{*}{\rotatebox{90}{HDR}} 
& HDR-NeRF~\cite{huang2022hdr}
&  22.84 & 0.9360 & 0.047
& 18.08 & 0.8729 & 0.066
& 22.01 & 0.9541 & 0.058
& 26.63 & 0.9523 & 0.046 \\
& HDR-GS~\cite{cai2024hdr}
&   22.98 & 0.8760 & 0.149
& 20.10 & 0.8444 & 0.202
& 7.94 & 0.4471 & 0.347
& 15.46 & 0.7648 & 0.099  \\
 & GaussHDR~\cite{liu2025gausshdr}
& 35.26 & 0.9664 & 0.027
& 36.50 & 0.9745 & 0.015
& 36.97 & 0.9675 & 0.015
& 34.42 & 0.9642 & 0.025 \\
 & GaussHDR$\dagger$~\cite{liu2025gausshdr}
&   39.26 & \cellcolor{red!30}0.9844 &  \cellcolor{yellow!30}0.007
& 37.52 & 0.9781 & 0.011
& 37.44 & 0.9704 & 0.012
& \cellcolor{red!30}35.82 & \cellcolor{red!30}0.9733 &  \cellcolor{yellow!30}0.014\\
&  Ours
&  \cellcolor{yellow!30}38.89 & 0.9826 & 0.008
& \cellcolor{yellow!30}37.73 &  \cellcolor{yellow!30}0.9784 &  \cellcolor{yellow!30}0.011
& \cellcolor{yellow!30}37.48 & \cellcolor{yellow!30}0.9705 & \cellcolor{yellow!30}0.011
& 34.53 & 0.9389 & 0.056\\
 & Ours$\dagger$
& \cellcolor{red!30}39.27 & \cellcolor{yellow!30}0.9842 & \cellcolor{red!30}0.006
& \cellcolor{red!30}37.85 & \cellcolor{red!30}0.9794 & \cellcolor{red!30}0.010
& \cellcolor{red!30}37.68 & \cellcolor{red!30}0.9718 & \cellcolor{red!30}0.011
& \cellcolor{yellow!30}35.74 & \cellcolor{yellow!30}0.9720 & \cellcolor{red!30}0.014 \\
\bottomrule
\end{tabular}}
\end{table*}

\begin{table*}[t]
\centering
  \caption{Per-scene quantitative comparisons on the HDR-NeRF-Syn~\cite{huang2022hdr} dataset under the exp1 exposure setting. For each scene, the best and second-best results are highlighted in \colorbox{red!30}{red} and \colorbox{yellow!30}{yellow}. LDR-OE and LDR-NE denote LDR results with exposure sets $\{t_1,t_3,t_5\}$ and $\{t_2,t_4\}$, respectively. }
\label{tab:hdrnerfsyn_per_sceneexp1}
\Large
\resizebox{1\linewidth}{!}{
  \begin{tabular}{c|l|ccc ccc ccc ccc}
\toprule
   \multicolumn{2}{c|}{\multirow{2}{*}{Method}}
  & \multicolumn{3}{c}{\textit{Bathroom}}
  & \multicolumn{3}{c}{\textit{Bear}}
  & \multicolumn{3}{c}{\textit{Chair}}
  & \multicolumn{3}{c}{\textit{Desk}} \\
  \cmidrule(lr){3-5}\cmidrule(lr){6-8}\cmidrule(lr){9-11}\cmidrule(lr){12-14}
  \multicolumn{2}{c|}{}& PSNR$\uparrow$ & SSIM$\uparrow$ & LPIPS$\downarrow$
& PSNR$\uparrow$ & SSIM$\uparrow$ & LPIPS$\downarrow$
& PSNR$\uparrow$ & SSIM$\uparrow$ & LPIPS$\downarrow$
& PSNR$\uparrow$ & SSIM$\uparrow$ & LPIPS$\downarrow$ \\
\midrule
   \multirow{6}{*}{\rotatebox{90}{LDR-OE}} 
& HDR-NeRF~\cite{huang2022hdr}
  &  33.97 & 0.9066 & 0.108
  & 43.81 & 0.9915 & 0.007
  & 34.31 & 0.9284 & 0.064
  & 38.26 & 0.9701 & 0.023 \\
  & HDR-GS~\cite{cai2024hdr}
  &   40.62 & 0.9734 & 0.011
  & 36.31 & 0.9745 & 0.020
  & 35.21 & 0.9610 & 0.022
  & 41.34 & 0.9847 & 0.006  \\
   & GaussHDR~\cite{liu2025gausshdr}
  & 41.02 & 0.9731 & 0.009
  & 44.02 & 0.9912 & 0.003
  & 37.01 & 0.9678 & 0.014
  &\cellcolor{yellow!30}42.46 & 0.9872 & 0.004 \\
   & GaussHDR$\dagger$~\cite{liu2025gausshdr}
  &  \cellcolor{yellow!30}42.08 & \cellcolor{red!30}0.9807 & \cellcolor{yellow!30}0.006
  & 45.40 & \cellcolor{red!30}0.9934 & 0.002
  & \cellcolor{red!30}37.65 & \cellcolor{yellow!30}0.9712 & \cellcolor{yellow!30}0.012
  & \cellcolor{red!30}43.03 & \cellcolor{red!30}0.9896 & \cellcolor{yellow!30}0.003 \\
 & Ours
 &  40.75 & 0.9740 & 0.009
 & \cellcolor{yellow!30}45.42 & 0.9933 & \cellcolor{yellow!30}0.002
 & 37.36 & 0.9695 & 0.013
 & 42.31 & 0.9884 & 0.004\\
  & Ours$\dagger$
 & \cellcolor{red!30}42.17 & \cellcolor{yellow!30}0.9801 & \cellcolor{red!30}0.006
 & \cellcolor{red!30}45.49 & \cellcolor{yellow!30}0.9933 & \cellcolor{red!30}0.002
 & \cellcolor{yellow!30}37.64 & \cellcolor{red!30}0.9712 & \cellcolor{red!30}0.012
 & 42.30 & \cellcolor{yellow!30}0.9895 & \cellcolor{red!30}0.003\\
  \midrule
   \multirow{6}{*}{\rotatebox{90}{LDR-NE}} 
& HDR-NeRF~\cite{huang2022hdr}
  &  33.29 & 0.9040 & 0.115
  & 42.35 & 0.9903 & 0.009
  & 33.51 & 0.9212 & 0.072
  & 36.47 & 0.9611 & 0.027\\
  & HDR-GS~\cite{cai2024hdr}
  &   34.99 & 0.9653 & 0.035
  & 12.58 & 0.7105 & 0.215
  & 27.47 & 0.9108 & 0.087
  & 33.47 & 0.9697 & 0.029\\
   & GaussHDR~\cite{liu2025gausshdr}
  & 40.81 & 0.9749 & 0.009
  & 43.48 & 0.9906 & 0.003
  & 36.54 & 0.9673 & 0.015
  & 42.36 & 0.9874 & 0.004 \\
   & GaussHDR$\dagger$~\cite{liu2025gausshdr}
  &  \cellcolor{yellow!30}41.70 & \cellcolor{red!30}0.9818 & \cellcolor{yellow!30}0.007
  & \cellcolor{yellow!30}44.43 & 0.9927 & 0.002
  & \cellcolor{yellow!30}37.14 & \cellcolor{yellow!30}0.9698 & \cellcolor{yellow!30}0.013
  & \cellcolor{red!30}42.76 & \cellcolor{yellow!30}0.9895 & \cellcolor{yellow!30}0.003 \\
 & Ours
 &  40.49 & 0.9754 & 0.009
 & 44.36 & \cellcolor{red!30}0.9927 & \cellcolor{yellow!30}0.002
 & 36.90 & 0.9686 & 0.014
 & 42.07 & 0.9888 & 0.004 \\
  & Ours$\dagger$
 & \cellcolor{red!30}41.85 & \cellcolor{yellow!30}0.9814 & \cellcolor{red!30}0.006
 & \cellcolor{red!30}44.50 & \cellcolor{yellow!30}0.9926 & \cellcolor{red!30}0.002
 & \cellcolor{red!30}37.28 & \cellcolor{red!30}0.9703 & \cellcolor{red!30}0.012
 & \cellcolor{yellow!30}42.10 & \cellcolor{red!30}0.9899 & \cellcolor{red!30}0.003\\
\midrule
   \multirow{6}{*}{\rotatebox{90}{HDR}} 
& HDR-NeRF~\cite{huang2022hdr}
  &  23.09 & 0.9558 & 0.090
  & 26.56 & 0.9649 & 0.023
  & 26.36 & 0.9554 & 0.057
  & 44.01 & 0.9932 & 0.007 \\
  & HDR-GS~\cite{cai2024hdr}
  &   20.01 & 0.8370 & 0.113
  & 14.01 & 0.7845 & 0.191
  & 12.64 & 0.0876 & 0.591
  & 22.79 & 0.4358 & 0.305  \\
   & GaussHDR~\cite{liu2025gausshdr}
  & 34.60 & 0.9401 & 0.025
  & 41.22 & 0.9855 & 0.007
  & 36.76 & 0.9621 & 0.020
  & 44.01 & 0.9931 & 0.006 \\
   & GaussHDR$\dagger$~\cite{liu2025gausshdr}
  &  \cellcolor{yellow!30}36.21 & \cellcolor{red!30}0.9522 & \cellcolor{yellow!30}0.017
  & \cellcolor{yellow!30}42.22 & \cellcolor{yellow!30}0.9881 & 0.005
  & \cellcolor{yellow!30}37.79 & \cellcolor{yellow!30}0.9659 & 0.017
  & 43.83 & 0.9932 & 0.005 \\
 &  Ours
 & 35.53 & 0.9436 & 0.019
 & 42.12 & \cellcolor{red!30}0.9881 & \cellcolor{yellow!30}0.005
 & 37.64 & 0.9653 & \cellcolor{yellow!30}0.017
 & \cellcolor{red!30}44.05 & \cellcolor{yellow!30}0.9932 & \cellcolor{yellow!30}0.005 \\
  & Ours$\dagger$
 & \cellcolor{red!30}36.31 & \cellcolor{yellow!30}0.9517 & \cellcolor{red!30}0.015
 & \cellcolor{red!30}42.31 & 0.9880 & \cellcolor{red!30}0.004
 & \cellcolor{red!30}37.92 & \cellcolor{red!30}0.9663 & \cellcolor{red!30}0.015
 & \cellcolor{yellow!30}43.96 & \cellcolor{red!30}0.9934 & \cellcolor{red!30}0.005\\
  \midrule  
   \multicolumn{2}{c|}{\multirow{2}{*}{ }}
  & \multicolumn{3}{c}{\textit{Diningroom}}
  & \multicolumn{3}{c}{\textit{Dog}}
  & \multicolumn{3}{c}{\textit{Sofa}}
  & \multicolumn{3}{c}{\textit{Sponza}} \\
  \cmidrule(lr){3-5}\cmidrule(lr){6-8}\cmidrule(lr){9-11}\cmidrule(lr){12-14}
  \multicolumn{2}{c|}{}& PSNR$\uparrow$ & SSIM$\uparrow$ & LPIPS$\downarrow$
& PSNR$\uparrow$ & SSIM$\uparrow$ & LPIPS$\downarrow$
& PSNR$\uparrow$ & SSIM$\uparrow$ & LPIPS$\downarrow$
& PSNR$\uparrow$ & SSIM$\uparrow$ & LPIPS$\downarrow$ \\
\midrule
   \multirow{6}{*}{\rotatebox{90}{LDR-OE}} 
& HDR-NeRF~\cite{huang2022hdr}
  &  43.29 & 0.9894 & 0.008
  & 39.49 & 0.9840 & 0.013
  & 39.71 & 0.9815 & 0.013
  & 36.61 & 0.9683 & 0.020 \\
  & HDR-GS~\cite{cai2024hdr}
  &   30.17 & 0.8914 & 0.130
  & 40.20 & 0.9888 & 0.007
  & 40.98 & 0.9888 & 0.006
  & 40.91 & 0.9880 & 0.005  \\
   & GaussHDR~\cite{liu2025gausshdr}
  & 40.47 & 0.9810 & 0.017
  & 43.17 & 0.9921 & 0.003
  & 43.50 & 0.9911 & 0.003
  & 41.73 & 0.9874 & 0.005 \\
   & GaussHDR$\dagger$~\cite{liu2025gausshdr}
  &  \cellcolor{red!30}45.21 & \cellcolor{red!30}0.9942 & \cellcolor{yellow!30}0.002
  & \cellcolor{red!30}43.76 & \cellcolor{red!30}0.9941 & \cellcolor{yellow!30}0.002
  &43.52 & 0.9919 & 0.003
  & \cellcolor{red!30}42.91 & \cellcolor{red!30}0.9911 & \cellcolor{yellow!30}0.004 \\
 & Ours
 &  44.02 & 0.9926 & 0.003
 & 43.12 & 0.9929 & 0.003
 & \cellcolor{yellow!30}43.56 & \cellcolor{yellow!30}0.9922 & \cellcolor{yellow!30}0.003
 & 39.39 & 0.9649 & 0.016\\
  & Ours$\dagger$
 & \cellcolor{yellow!30}44.96 & \cellcolor{yellow!30}0.9941 & \cellcolor{red!30}0.002
 & \cellcolor{yellow!30}43.25 & \cellcolor{yellow!30}0.9939 & \cellcolor{red!30}0.002
 & \cellcolor{red!30}44.08 & \cellcolor{red!30}0.9932 & \cellcolor{red!30}0.002
 & \cellcolor{yellow!30}42.48 & \cellcolor{yellow!30}0.9896 & \cellcolor{red!30}0.004\\
  \midrule
   \multirow{6}{*}{\rotatebox{90}{LDR-NE}} 
& HDR-NeRF~\cite{huang2022hdr}
  &  42.36 & 0.9892 & 0.008
  & 38.46 & 0.9822 & 0.014
  & 38.56 & 0.9802 & 0.013
  & 36.04 & 0.9683 & 0.021 \\
  & HDR-GS~\cite{cai2024hdr}
  &   25.62 & 0.8598 & 0.175
  & 15.44 & 0.7160 & 0.305
  & 18.00 & 0.7981 & 0.229
  & 35.75 & 0.9810 & 0.013 \\
   & GaussHDR~\cite{liu2025gausshdr}
  &  39.62 & 0.9798 & 0.020
  & \cellcolor{yellow!30}42.40 & 0.9916 & 0.004
  & 42.90 & 0.9913 & 0.003
  & 41.56 & 0.9890 & 0.005\\
   & GaussHDR$\dagger$~\cite{liu2025gausshdr}
  &  43.33 & \cellcolor{yellow!30}0.9934 & 0.003
  & 42.12 & \cellcolor{yellow!30}0.9933 & 0.003
  & 42.04 & 0.9914 & 0.003
  & \cellcolor{red!30}42.60 & \cellcolor{red!30}0.9916 & \cellcolor{yellow!30}0.004\\
 & Ours
 & \cellcolor{yellow!30}43.36 & 0.9925 & 0.003
 & 42.15 & 0.9925 & \cellcolor{yellow!30}0.003
 & \cellcolor{yellow!30}42.96 & \cellcolor{yellow!30}0.9923 & \cellcolor{yellow!30}0.002
 & 39.46 & 0.9707 & 0.016 \\
  & Ours$\dagger$
 & \cellcolor{red!30}43.58 & \cellcolor{red!30}0.9938 & \cellcolor{red!30}0.002
 & \cellcolor{red!30}42.40 & \cellcolor{red!30}0.9938 & \cellcolor{red!30}0.002
 & \cellcolor{red!30}43.49 & \cellcolor{red!30}0.9935 & \cellcolor{red!30}0.002
 & \cellcolor{yellow!30}41.83 & \cellcolor{yellow!30}0.9906 & \cellcolor{red!30}0.004\\
\midrule
   \multirow{6}{*}{\rotatebox{90}{HDR}} 
& HDR-NeRF~\cite{huang2022hdr}
  &  30.17 & 0.9873 & 0.017
  & 22.84 & 0.9359 & 0.047
  & 18.16 & 0.8749 & 0.063
  & 21.68 & 0.9514 & 0.062 \\
  & HDR-GS~\cite{cai2024hdr}
  &   20.88 & 0.8194 & 0.205
  & 20.16 & 0.8476 & 0.185
  & 9.85 & 0.6003 & 0.296
  & 11.10 & 0.4262 & 0.234  \\
   & GaussHDR~\cite{liu2025gausshdr}
  & 35.49 & 0.9666 & 0.029
  & 36.30 & 0.9734 & 0.016
  & 36.81 & 0.9666 & 0.016
  & 34.36 & 0.9655 & 0.021 \\
   & GaussHDR$\dagger$~\cite{liu2025gausshdr}
  &  \cellcolor{yellow!30}38.61 & \cellcolor{yellow!30}0.9832 & 0.008
  & \cellcolor{yellow!30}37.51 & \cellcolor{yellow!30}0.9780 &  0.012
  &  37.13 & 0.9691 & 0.013
  & \cellcolor{red!30}35.64 & \cellcolor{red!30}0.9720 & \cellcolor{yellow!30}0.014 \\
 &  Ours
 &  38.57 & 0.9817 & \cellcolor{yellow!30}0.008
 &  37.37 &  0.9774 & \cellcolor{yellow!30}0.011
 & \cellcolor{yellow!30}37.22 & \cellcolor{yellow!30}0.9693 & \cellcolor{yellow!30}0.011
 & 34.26 & 0.9418 & 0.030\\
  & Ours$\dagger$
 & \cellcolor{red!30}39.04 & \cellcolor{red!30}0.9837 & \cellcolor{red!30}0.006
 & \cellcolor{red!30}37.68 & \cellcolor{red!30}0.9790 & \cellcolor{red!30}0.010
 & \cellcolor{red!30}37.46 & \cellcolor{red!30}0.9710 & \cellcolor{red!30}0.011
 & \cellcolor{yellow!30}35.60 & \cellcolor{yellow!30}0.9718 & \cellcolor{red!30}0.014\\
\bottomrule
\end{tabular}}
\end{table*}

\section{Per-scene Quantitative Results}\label{sec:quanti} 
We report per-scene quantitative results on HDR-NeRF-Real~\cite{huang2022hdr}, HDR-Plenoxels-Real~\cite{jun2022hdr}, and HDR-NeRF-Syn~\cite{huang2022hdr} in \Cref{tab:hdrnerfreal_per_scene}, \Cref{tab:hdrplenreal_per_scene}, \Cref{tab:hdrnerfsyn_per_sceneexp3}, and \Cref{tab:hdrnerfsyn_per_sceneexp1}. On HDR-NeRF-Real, Ours$\dagger$ achieves leading performance, obtaining the best scores on most scenes, while the 3DGS variant (Ours) also shows competitive results and delivers the best on several scenes. Notably, compared with the strong GaussHDR variants~\cite{liu2025gausshdr}, our method delivers higher rendering speed (e.g., Ours$\dagger$ is $1.37\times$ faster than GaussHDR$\dagger$) and lower memory usage during training. On HDR-Plenoxels-Real, Ours$\dagger$ again shows overall superior performance over all compared methods. On the synthetic HDR-NeRF-Syn dataset, Ours$\dagger$ consistently outperforms baselines across both exp3 and exp1 exposure settings, achieving the best results on most scenes. This highlights our superiority in reconstructing HDR details, as well as the tone mapping ability to faithfully preserve standard sensor observations.

\vspace{-2pt}\section{More Qualitative Results}\label{sec:qualitative}
\begin{figure*}[t]
\centering
\vspace{-0.01in}\includegraphics[width=1\linewidth, clip=true, trim=20pt 30pt 20pt 18pt]{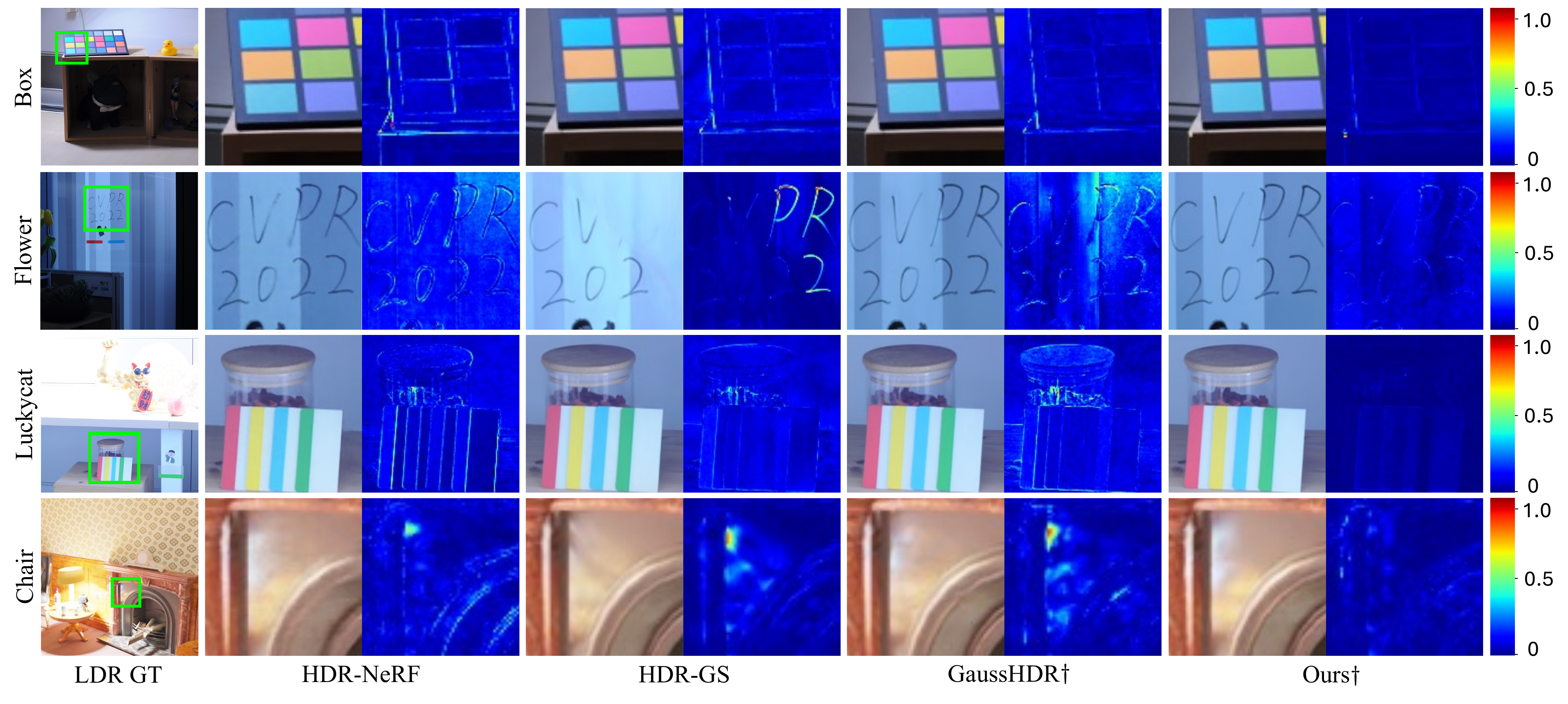} \vspace{-0.19in}
\caption{Qualitative results on LDR views, where residual maps between the rendered results and ground truth are visualized to highlight the difference. HDR-GS struggles to balance between dynamic-range content reconstruction and fine-structure preservation(\eg, characters in the highlight region of the 2nd row are missing), whereas our approach delivers accurate dynamic-range estimations and sharper details.
}  
\label{fig:ldr_supp}
\end{figure*}

\begin{figure*}[t]
\centering
\includegraphics[width=1\linewidth, clip=true, trim=20pt 30pt 20pt 18pt]{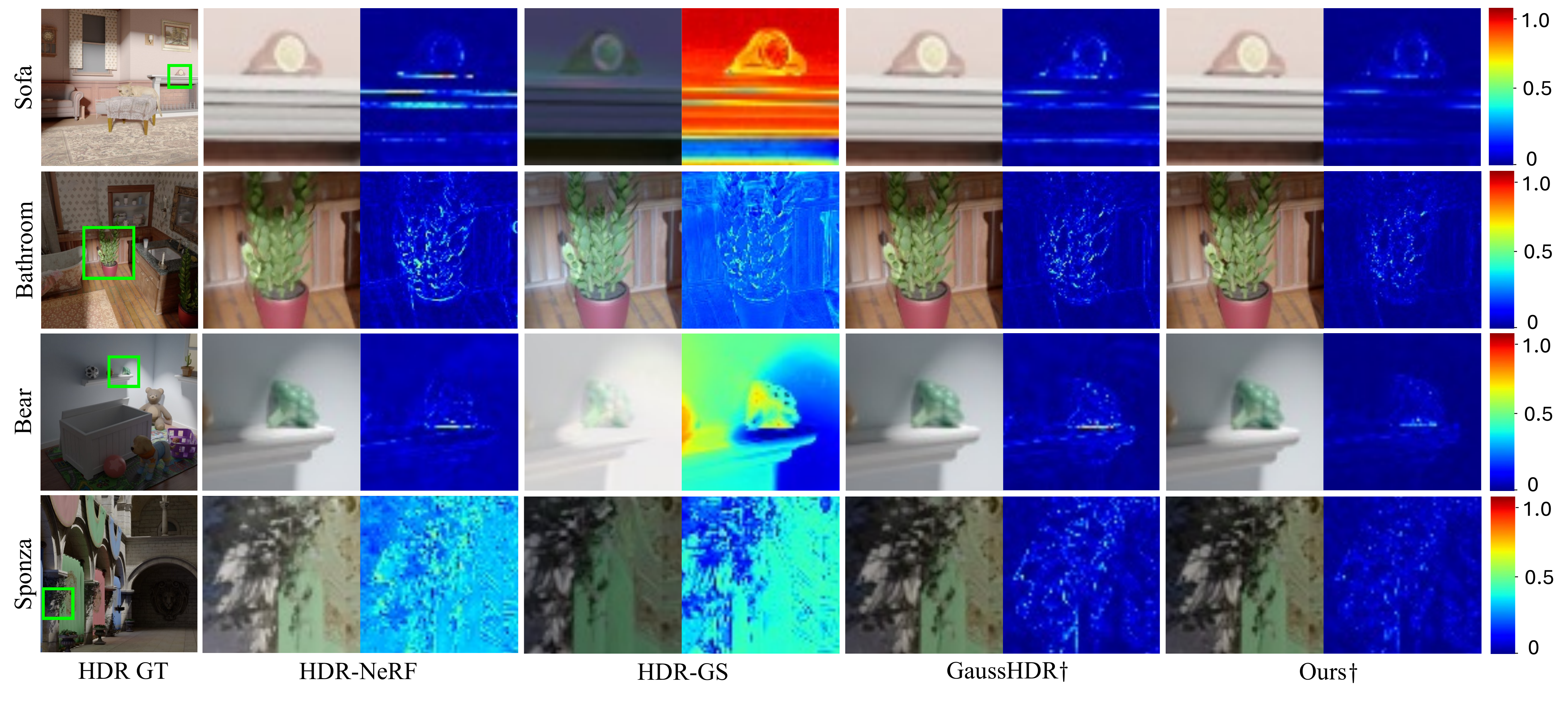} 
\caption{Qualitative results on HDR views, where residual maps between the rendered results and ground truth are visualized to highlight the differences. HDR-NeRF and HDR-GS fail to reach the target illumination level due to the absence of HDR ground truth during training, whereas our results yield accurate dynamic-range estimations and sharper reconstructions (\eg, leaf textures in the 4th row).}  
\label{fig:hdr_supp}
\end{figure*}

\vspace{-2pt}
We provide additional qualitative LDR results in \Cref{fig:ldr_supp} and HDR results in \Cref{fig:hdr_supp}. For each method, we also visualize the residual map with respect to the ground truth to intuitively show the difference (from blue to red, higher values indicate larger errors).  For LDR results, existing methods struggle to balance reconstructing dynamic range content and preserving fine structures (\eg, the texture details of HDR-GS~\cite{cai2024hdr} at the highlight region are missing). In contrast, our method preserves both dynamic-range detail and local structures, and even reflects environment-conditioned appearance changes (\eg, the reflection of the fireplace), which we attribute to the complementary IE and GI branches that decouple exposure handling and illumination modeling. For HDR results, HDR-NeRF~\cite{huang2022hdr} and HDR-GS often fail to reach the correct illumination level due to the lack of HDR ground truth during training, while our approach produces more accurate dynamic-range estimations and sharper reconstructions (\eg, leaf textures in the 4th row), demonstrating the ability to restore dynamic ranges and keep structural details.

\vspace{-3pt}\section{Experimental Settings}\label{sec:setting}\vspace{-3pt}
\noindent\textbf{Datasets and settings.} 
For the HDR-Plenoxels-Real~\cite{jun2022hdr} dataset, LDR-NE ($\{t_2,t_4\}$) views are not provided for the \textit{Character} and \textit{Coffee} scenes. Therefore, for LDR-NE results on HDR-Plenoxels-Real, we report the average performance over the remaining \textit{Desk} and \textit{Plant} scenes. For HDR-NeRF-Real~\cite{huang2022hdr}, HDR-Plenoxels-Real~\cite{jun2022hdr}, and the synthetic HDR-NeRF-Syn~\cite{huang2022hdr}, we follow the previous method~\cite{liu2025gausshdr} to downsample the original resolutions by factors of $1/4$, $1/6$, and $1/2$, yielding training resolutions of $804 \times 534$, $992 \times 746$, and $400 \times 400$, respectively. 

\noindent\textbf{Implementation details.} 
During training, we adopt the Adam optimizer~\cite{kingma2014adam} to optimize the Gaussian parameters, the radiance composer $g$, the illumination modulator $\varphi$, and the tone mapper $f$. The initial learning rates for $g$ and $\varphi$ are set to $6\times 10^{-5}$, and the learning rate for the tone mapper is set to $2\times 10^{-4}$. All learning rates are scheduled with a cosine annealing scheme~\cite{loshchilov2016sgdr}. The densification threshold $\tau_p$ is set to $2\times 10^{-4}$.

\section{Discussions}\label{sec:discussion}
During training, we observe that LDR views captured in low-light conditions may contain significant sensor noise. Fitting view-specific noise can introduce floaters/thin Gaussians. This can be mitigated by applying a noise-prior guided regularization on per-view reconstruction to discourage fitting high-frequency noise. Meanwhile, stronger geometry backbones (\eg, the adopted Scaffold-GS~\cite{lu2024scaffold}) can further suppress floaters by constraining Gaussians around anchor points.

{
    \small
    \bibliographystyle{ieeenat_fullname}
    \bibliography{main}
}


\end{document}